\documentclass[10pt,twocolumn,letterpaper]{article}

\usepackage{cvpr}
\usepackage{times}
\usepackage{epsfig}
\usepackage{graphicx}
\usepackage{amsmath}
\usepackage{mathtools}
\usepackage{amssymb}
\usepackage{caption}
\usepackage{epsfig}
\usepackage{epstopdf}
\usepackage{graphicx}
\usepackage{subfig}
\usepackage{float}
\usepackage{bbold}
\usepackage{multirow}
\usepackage{hhline}
\usepackage{array}

\newcommand{\comment}[1]{}

\newcommand{\pfrmk}[1]{\textcolor{red}{\bf pf: #1}}
\newcommand{\pf}[1]{{\color{red} #1}}
\newcommand{\amrmk}[1]{\textcolor{blue}{\bf am: #1}}
\newcommand{\am}[1]{{\color{blue} #1}}

\newcommand{\basket}[1]{Basket-#1}
\newcommand{\volley}[1]{Volley-#1}
\newcommand{\soccer}[1]{Soccer-#1}

\newcommand{\our}{{\bf OUR}}
\newcommand{\ournp}{{\bf OUR-No-Physics}}
\newcommand{\ourns}{{\bf OUR-Two-States}}
\newcommand{\inter}{{\bf InterTrack}}
\newcommand{\ksp}{{\bf KSP}}
\newcommand{\maxdet}{{\bf MaxDetection}}
\newcommand{\comptrack}{{\bf CompTrack}}
\newcommand{\ransac}{{\bf RANSAC}}
\newcommand{\growth}{{\bf Growth}}
\newcommand{\fos}{{\bf FoS}}

\usepackage[pagebackref=true,breaklinks=true,letterpaper=true,colorlinks,bookmarks=false]{hyperref}

\cvprfinalcopy 

\def\cvprPaperID{1862} 

\ifcvprfinal\fi
\begin{document}

\title{What Players do with the Ball: A Physically Constrained Interaction Modeling}
\author{Andrii Maksai \qquad Xinchao Wang \qquad Pascal Fua\\
Computer Vision Laboratory, \'Ecole Polytechnique F\'ed\'erale de Lausanne (EPFL)\\
{\tt\small \{andrii.maksai, xinchao.wang, pascal.fua\}@epfl.ch}
}

\maketitle

\begin{abstract}
Tracking the ball is critical for  video-based analysis of team sports. However,
it is difficult,  especially in low-resolution images, due to  the small size of
the ball, its  speed that creates motion  blur, and its often  being occluded by
players.

In this  paper, we  propose a  generic and principled  approach to  modeling the
interaction between  the ball  and the players  while also  imposing appropriate
physical constraints on the ball's trajectory.

We show  that our approach, formulated  in terms of  a Mixed Integer  Program, is
more robust and more accurate  than several state-of-the-art
approaches on real-life volleyball, basketball, and soccer sequences.

\end{abstract}


\section{Introduction}

Tracking the ball  accurately is critically important to  analyze and understand
the action in  sports ranging from tennis to soccer,  basketball, volleyball, to
name  but a  few.  While commercial  video-based systems  exist  for the  first,
automation remains elusive  for the others. This is largely  attributable to the
interaction between  the ball and the  players, which often results  in the ball
being either  hard to detect because  someone is handling it  or even completely
hidden  from view.  Furthermore, since  the players  often kick  it or  throw it
in  ways  designed  to  surprise  their opponents,  its  trajectory  is  largely
unpredictable.

\begin{figure*}[t]
\begin{center}
\hspace{-1pt}
\begin{tabular}{ccc}
   \hspace{-0.2cm}\includegraphics[height=3.6cm]{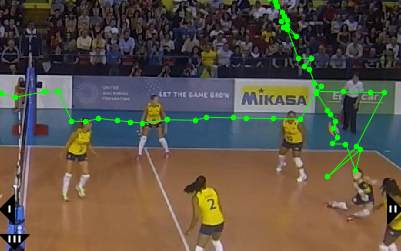} &
   \hspace{-0.2cm}\includegraphics[height=3.6cm]{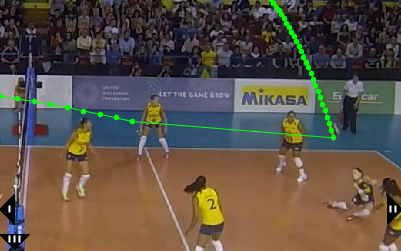} &
   \hspace{-0.2cm}\includegraphics[height=3.6cm]{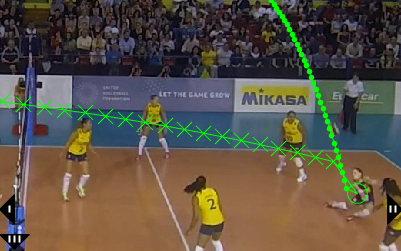} 
\\   
   \hspace{-0.2cm} (a) &
   \hspace{-0.2cm} (b) &
   \hspace{-0.2cm} (c)
\end{tabular}
\end{center}
\vspace{-6mm}

\caption{Importance  of   simultaneously  modeling  interactions   and  imposing
physical constraints.  For most of  this 70-frame volleyball  sequence depicting
the ball crossing the net and being bumped by a defending player and viewed by 3
cameras,  the defending  player is  on  the ground.  As  a result,  she was  not
detected by the  person detector we use~\cite{Fleuret08a} because  it only finds
people standing  up. Furthermore,  while the  ball was near  the player,  it was
occluded in the views  of 2 of the 3 cameras, and, therefore,  not detected as a
3D object.  {\bf (a)} Tracking the  players and the ball  simultaneously without
imposing motion constraints as  in~\cite{Wang14b} produces physically impossible
trajectories. {\bf (b)} Imposing motion constraints but tracking the players and
the ball  separately as in~\cite{Parisot15}  does not properly capture  the ball
and player interaction. {\bf (c)} Our  approach to both imposing constraints and
modeling the interaction  gives a better overall result. The  crosses denote the
fact that  the ball is  in the  ``strike'' state until  being bumped and  in the
``flying''  one after  that. Transitions  between these  states can  only result
from  interacting  with  a  player,  which  encourages  the  optimizer  to  find
one  in spite  of  the weak  evidence. Best  viewed  in color.}  \vspace{-2.0mm}
\label{fig:motivation} \end{figure*}

There  is   a  substantial   body  of  literature   about  dealing   with  these
issues,   but  almost   always  using   heuristics  that   are  specific   to  a
particular sport  such as soccer~\cite{Zhang08b},  volleyball~\cite{Gomez14}, or
basketball~\cite{Chen09a}.\comment{\pfrmk{We'll need to argue that we outperform
them someplace.}  \amrmk{For the  first, unfortunately,  there is  not available
code.  For  the last  two,  I  hope to  be  able  to use  part  of  the code  of
~\cite{Gomez14}, which  implements exactly the approach  of ~\cite{Chen09a}, but
for another sport. This  way, we would be able to claim that  we are better than
both  of them.}}  A  few  more generic  approaches  explicitly  account for  the
interaction between the players and  the ball~\cite{Wang14b} while others impose
physics-based constraints  on ball motion~\cite{Parisot15}. However,  neither of
these things  alone suffices  in difficult  cases, such as  the one  depicted by
Fig.~\ref{fig:motivation}.

In this paper,  we, therefore, introduce an approach  to simultaneously accounting
for ball/player interactions and imposing appropriate physics-based constraints.
Our  approach  is generic  and  applicable  to  many  team sports.  It  involves
formulating the ball tracking problem in  terms of a Mixed Integer Program (MIP)
in which we account  for the motion of both the players and  the ball as well as
the fact the  ball moves differently and has different  visibility properties in
flight,  in  possession  of  a  player,  or while  rolling  on  the  ground.  We
model the  ball locations  in $\mathbb{R}^3$ and  impose first  and second-order
constraints where  appropriate. The resulting  MIP describes the  ball behaviour
better  than previous  approaches~\cite{Wang14b,Parisot15}  and yields  superior
performance, both  in terms of  tracking accuracy and robustness  to occlusions.
Fig.~\ref{fig:motivation}(c) depicts  the improvement resulting from  doing this
rather than  only modeling the  interactions or only imposing  the physics-based
constraints.


In  short,  our  contribution  is   a  principled  and  generic  formulation  of
the  ball   tracking  problem   and  related   physical  constraints   in  terms
of   a  MIP.   We  will   demonstrate  that   it  outperforms   state-of-the-art
approaches~\cite{Wang14a,Wang14b,Parisot15,Gomez14}  in soccer,  volleyball, and
basketball.

\comment{\pfrmk{Will  you have soccer  results as well?  Apparently,  we have
  only  one baseline  that  is not  us, i.e.~\cite{Parisot15}.   \cite{Zhang12a}
  doesn't really  count since it  was never  designed for this.}   \amrmk{Yes, I
  will upload  the results  of experiments  with football in  a couple  of days.
  There is also a work of~\cite{Ren08}, with which we can make strong parallels,
  as  one of  our baselines  does something  quite similar-first  tracking, then
  classification   of   ball   states   -but  their   code   is   not   publicly
  available. Also~\cite{Gomez14}  has a  5k line code  without the  sample data,
  examples of usage  or comments. I will  hopefully be able to  extract the ball
  tracking part from it.}}

\comment{For more than half of this short sequence, the player catching and then
  throwing back the ball  was on the ground. As a result  the player detector we
  use   failed   to   find   her   because  it   models   people   as   vertical
  cylinders~\cite{Fleuret08a}.  Furthermore, while the ball was near the player,
  it was occluded in the views of cameras 1 and 3, and, therefore, not detected.
  A model  that does not consider  interactions (Fig.~\ref{fig:motivation}.(b)),
  is able  to track  the ball for  the parts of  ballistic trajectories,  but it
  provides  implausible solution  in  the middle.   An  approach that  considers
  interactions     but      has     no     notion     of      physical     model
  (Fig.~\ref{fig:motivation}.(a)), also chooses a solution that does not require
  presence of an undetected person,  building physically impossible solution for
  the frames where the ball is not detected. Physical model of our solution does
  not allow  the acceleration change for  the flying ball and  therefore reasons
  correctly about the presence of a  person. Additional cues are provided by the
  ball state classifiers and leared priors: the most likely state of the ball in
  the first part of the trajectory is  `Strike', while in the second part of the
  trajectory it  is `Flying'.  The  probability of transition between  these two
  states is zero, as it requires the ball to be possessed in between.}

\comment{Our approach tracks the players and the ball and assigns a state to the
ball  at each  time instant  simultaneously. By  exploiting multiple  sources of
information, our  system finds the  trajectory of the  ball and the  sequence of
states  that  is  highly  likely  with  respect to  the  location  of  the  ball
(\eg  in basketball  higher  location of  the  ball above  the  ground is  often
correlated with  the locations  near the  basket and the  state of  being shot),
physical constraints,  and visibility  (\eg it  is often likely  not to  see the
basketball during  dribling). Furthermore, our novel  problem formulation allows
to effectively use detections, that are discrete by nature, to form a trajectory
in continuous  space, while  preserving global optimality  of the  solution. Our
contributions  are  three-fold: \begin{itemize}  \item  Novel  scheme of  object
tracking  and  state  estimation  under  physical  constraints  that  allow  for
globally optimal solution without the  discretization of the search space; \item
Domain-independent  approach  for  tracking  the ball  that  requires  only  the
training data  annotated with  the state of  the ball but  does not  include any
domain-specific assumptions  in the optimization; \item  Ball tracking framework
that compares favourably to several baselines and state-of-the-art approaches on
real-life basketball and volleyball datasets. \end{itemize}

The rest of the paper is organized as follows. Related work is discussed in
Section~\ref{sec:related}. In Section~\ref{sec:problem}, we formulate the
problem of object tracking and state estimation under physical constraints. In
Section~\ref{sec:learning} we describe the specific form of factors we use and
the process of learning them. Experimental results are shown in
Section~\ref{sec:experiments}.
}

\comment{Precise  ball tracking  in sports  is one  of the  starting points  for
automatic game  analysis, results of which  are of huge importance  to referees,
coaches and athletes. Ball trajectories are  often used for understanding of the
higher-level semantics of  the game. With the advances in  computer vision, many
applications for automated detection, tracking  of ball and players and analysis
of the  game have been  developed. While detection  and tracking of  players can
often be  done reliably,  same can  not be said  for the  ball tracking  in many
sports. Rare  occlusions and discriminative  colour features of the  tennis ball
allow  it to  be tracked  with great  precision, but  for team  sports, such  as
volleyball,  basketball or  soccer, the  situation is  very different.  Multiple
occlusions by the players, fast speed  and often unpredictable trajectory of the
ball make  tracking the  ball much  more difficult  compared to  typical generic
object  tracking. As  a result,  most applications  concentrate on  one specific
sport or  even setup,  incorporating as  much domain  knowledge as  possible for
better tracking, sacrificing  the ability to apply the system  for other sports.
Having universal  solutions for tracking  people but using  domain-specific ball
trackers often means that tracking the ball, tracking the players and extracting
semantic information (\eg passes between players) are all done separately. Error
on any stage of  this process propagates further and it is  hard to recover from
it. }


\section{Related work}
\label{sec:related}

\comment{ Some approaches to game  understanding exploit for adversarial behavior
  discovery~\cite{Lucey13},  role assignment~\cite{Lan12},  activity recognition
  handball~\cite{Direkoglu12,Gupta09},    tracking   players~\cite{Liu13},    or
  discovering  regions of  interest~\cite{Kim12}.   Many  such approaches  could
  benefit from  tracking the  ball~\cite{Direkoglu12}. Others~\cite{Li09c,Li09d}
  use manually annotated  ball/puck data for game element  recognition in hockey
  and American football. The rest rely on tracking the ball.  }

While     there    are     approaches     to     game    understanding,     such
as~\cite{Lan12,Liu13,Lucey13,Gupta09,Direkoglu12,Kim12},   which  rely   on  the
structured nature of the data without  any explicit reference to the location of
the ball,  most others either  take advantages of  knowing the ball  position or
would benefit from being able  to~\cite{Direkoglu12}. However, while the problem
of automated ball tracking  can be considered as solved for  some sports such as
tennis or golf, it remains difficult  for team sports. This is particularly true
when the image resolution  is too low to reliably detect  the ball in individual
frames  in  spite  of  frequent  occlusions.  \comment{\pfrmk{Last  sentence  is
right.}}

Current approaches to detecting and tracking  can be roughly classified as those
that build  physically plausible  trajectory segments  on the  basis of  sets of
consecutive  detections  and  those  that  find  a  more  global  trajectory  by
minimizing an objective function. We briefly review both kinds below.

\subsection{Fitting Tracjectory Segments}

Many      ball-tracking      approaches     for      soccer~\cite{Ohno00,Leo08},
basketball~\cite{Chen09a},   and  volleyball~\cite{Chen07,Gomez14,Chakraborty13}
start with a set of successive detections that obey a physical model. They
then   greedily   extend   them   and  terminate   growth   based   on   various
heuristics. \comment{~\cite{Leo08,Chen07,Chen09a,Gomez14} grow the trajectory by
  finding the next candidate that fits the model, while~\cite{Chakraborty13} use
  Kalman filter  for the same  purpose. ~\cite{Leo08} use intersections  of ball
  and     players    trajectories     to     identify    interaction     events,
  while~\cite{Chen07,Gomez14} grow the neighbouring  pairs of trajectories until
  intersection  to identify  events of  ball-player contact.}   In~\cite{Ren08},
Canny-like hysteresis  is used to  select candidates above a  certain confidence
level and link them to  already hypothesized trajectories. Very recently, RANSAC
has been used to segment ballistic  trajectories of basketball shots towards the
basket~\cite{Parisot15}.   These   approaches  often  rely  heavily   on  domain
knowledge,  such  as audio  cues  to  detect  ball hits~\cite{Chen07}  or  model
parameters adapted to specific sports~\cite{Chakraborty13,Chen09a}.

While  effective when  the initial  ball detections  are sufficiently  reliable,
these methods tend to suffer from their  greedy nature when the quality of these
detections  decreases. We  will  show this  by comparing  our  results to  those
of~\cite{Gomez14,Parisot15}, for which  the code is publicly  available and have
been shown to be a good representatives of this set of methods.

\comment{\pfrmk{Can  you say  why you  chose those?}\amrmk{They
are  publicly  available  &  do  not require  external  knowledge  (audio  cues,
etc.),~\cite{Parisot15}  shows  more promise  as  it  doesn't detections  to  be
adjacent,~\cite{Gomez14} reimplements  the approach  which claimed to  have over
90\% accuracy}.}

\subsection{Global Energy Minimization}

One way to increase robustness is to  seek the ball trajectory as the minimum of
a global  objective function.  It  often includes high-level  semantic knowledge
such  as players'  locations~\cite{Zhu07a,Zhang08b,Wang14a}, state  of the  game
based on  ball location,  velocity and  acceleration~\cite{Zhang08b,Zhu07a}, or
goal events~\cite{Zhu07a}.

In~\cite{Wang14b,Wang15},   the  players   {\it  and}   the  ball   are  tracked
simultaneously and ball possession is  explicitly modeled. However, the tracking
is performed on a discretized  grid and without physics-based constraints, which
results in  reduced accuracy.  It has  nevertheless been shown  to work  well on
soccer and  basketball data. We  selected it as  our baseline to  represent this
class of methods, because of its state-of-the-art results and publicly available
implementation. \comment{as a  representative of this class Since we  know of no
other global  optimization technique  that takes advantage  of both  context and
physics-based constraints, we use it as one of our baselines.}

\comment{
\subparagraph{Approaches without  the ball}  concentrate on  detecting
interactions without relying on localizing the ball.

\cite{Poiesi10,Kim12} propose a detector-less approach  based on the motion flow
and stochastic  vector field of motion  tendencies, accordingly,~\cite{Poiesi10}
links points  of convergence of the  motion flow into the  trajectories by using
the Kalman  filter, while~\cite{Kim12} generate  regions of interest but  do not
report how  often the  ball is located  within it.

For~\cite{Poiesi10}, authors
report a high accuracy of 82\%, but  are unable to predict an actual location of
the ball within the area of flow convergence.

Other   approaches  concentrate   on   game  element   recognition~\cite{Li09d},
role     assignment~\cite{Lan12}     in     hockey,     activity     recognition
in     handball~\cite{Direkoglu12}     and    baseball~\cite{Gupta09},     plays
recognition~\cite{Li09c}   in  American   football.   In  all   of  the   above,
authors   either   manually  detect   the   ball/puck   or  use   ground   truth
data~\cite{Li09d,Li09c}, or  don't use the  location in the  problem formulation
and mention that their approach could benefit from it~\cite{Direkoglu12}.

}
\comment{
provide more  principled  way  of
finding a  trajectory optimal with respect  to a certain cost  function. In this
case cost function often  includes~ the    use  it  to classify  the  state  of  the
football game.~\cite{Zhang08b} train  an Adaboost classifier to filter  a set of
candidate trajectories obtained by using  a particle filter with the first-order
linear model. The final ball trajectory is generated using both people detection
and candidate ball trajectories, but,  unfortunately, no specific details of the
ball tracking  formulation are given, and  evaluation is done only  on the short
sequences. \cite{Zhu07a} intialize the tracking by Viterbi algorithm to link the
detections, and  continue by  using the  SVR particle  filter, but  use external
information sources to detect the goal events, which are of main interest to the
authors.  \cite{Wang14a}  concentrate  on  tracking  the ball  while  it  is  in
possession, by predicting and exploiting high  level state of the game. Approach
can be  viewed as simultaneous ball  tracking and game state  estimation, yet it
relies on correct tracking of the ball while it is not possesseed.
}

\comment{
These approaches  tend to  rely heavily on  domain knowledge,  either explicitly
(\cite{Chen07}   use  audio   cues  to   detect  the   moment  of   hitting  the
ball;~\cite{Chakraborty13,Chen09a} bound the possible parameters of the physical
model  specific to  a type  of sport;~\cite{Ren08}  create classifiers  based on
domain knowledge;~\cite{Gomez14,Parisot15}  assume the  ball always  undergoes a
ballistic motion) or implicitly, by proposing an approach for a specific sport.
}

\comment{,  one
of  \textit{flying,  rolling,  in\_possession, out\_of\_play}.  Classifier  uses
information  about  the location  of  the  ball,  likelihood of  the  detection,
distance to  the nearest  player and the  distance from the  edge of  the field.
Depending on  the selected state, trajectory  in the missed frames  is estimated
using a physical model specific for a particular type of motion. This is similar
to the  classifier that we are  using, but in our  work classifier is used  as a
part of  the globally optimal  tracking process. Furthermore,  authors associate
\textit{in\_possession} with a low detector output,  use it as an initializer to
other stages, and  focus on tracking the  visible ball, while our  work uses the
state as an ``equal partner''.}
\comment{Due to the local nature of  the optimization in the methods, they often
perform well only when a ball is visible and follows an easy ballistic or linear
trajectory:~\cite{Parisot15} reports the drop of  tracking accuracy from 70\% to
33\% when trying  to track in the whole field  area;~\cite{Gomez14} report below
60\% accuracy  when using the  method of~\cite{Chen07}, while the  original work
claims to have over  90\% of accuracy, a difference that  could be attributed to
the fact that~\cite{Chen07} used external cues  to detect beginning of the play.
}

\comment{
\subparagraph{Trajectory growth and  fitting} selects a trajectory  based on the
set of  the detections and the  appropriate physical model. 

One of the  earliest works\cite{Ohno00} finds the trajectory of  the football by
minimizing the  fitting error of  ballistic trajectory. No actual  evaluation of
tracking accuracy was done.

\cite{Leo08} track the football ball  by identifying candidate trajectories that:
join  detections  that  are  collinear  when  projected  on  the  ground  plane.
Trajectories are  extended greedily  while the  next detection  is found  in the
centain small time window. Later intersections  of the ball trajectories and the
players are used to identify the interation events.

\cite{Ren08} first generate  trajectories in each of the camera  views, by using
Canny-like hysteresis  to select candidates  above a certain threshold  or those
connected  to  already  selected  candidates.  However,  only  the  most  likely
candidate from each single view is passed further on to estimate the 3D location
of the football.  Heuristic-based classifiers are used to estimate  the state of
the  ball,  one  of  \textit{flying,  rolling,  in\_possession,  out\_of\_play}.
Classifiers use  information about the location  of the ball, likelihood  of the
detection, distance to the nearest player and  the distance from the edge of the
field.  Depending on  the selected  state, trajectory  in the  missed frames  is
estimated using a physical model specific  for a particular type of motion. This
work associates ``in possession'' with a low detector output. This state is used
as an initializer to  other stages and the focus is on the  stages when the ball
is more visible.

\cite{Chen09a}  link  basket  ball  candidates  on  the  image  plane  and  grow
trajectories based on  the physical model with  a maximum of 5  frames miss. The
set of candidates  is filtered through a thorough analysis  of possible physical
characteristics of  the trajectory based on  the height of the  basket and other
domain knowledge, and sorted based on heuristics taking into account the length,
fitting error and ratio of  isolated candidates. Afterwards, 2D trajectories are
mapped into  3D trajectories with the  aim of estimating the  shooting location.
Authors also proposed  variations of such approach for  volleyball, baseball and
football. Proposed  solutions heavily rely  on domain knowledge (\eg  audio cues
are used to identify  the moment when the player hits the  ball or the beginning
of  the game,  ballistic trajectories  of  the ball  are filtered  based on  the
knowledge of  the physical properties  of the  given ball, etc.).  While authors
claim  to have  typically over  90\% of  tracking accuracy  in a  small vicinity
of  the  ball,  no  publicly  available  implementation  of  their  approach  is
given.  However, the  comparison of  their method  to some  other methods,  made
by~\cite{Gomez14},  while  identified  its  superiority,  reveals  the  tracking
accuracy of below 60\%. Latter authors also provide their implementation.

\cite{Chakraborty13} start from close pairs of volleyball candidates in nearby
frames. They use Kalman filter to generate trajectory candidate, which is
terminated based on the number of missed candidates in the trajectory. Afterwards,
candidates are selected starting from the longest one, subject to having
appropriate physical parameters. Gaps are filled by interpolating and
extrapolating the trajectory, but only up to 5 frames. This information is
afterwards used to classify shots based on the geometric properties of
trajectories, specified by the authors.

More  recent work~\cite{Parisot15}  uses RANSAC  on  the 3D  candidates to  form
ballistic  trajectories in  basketball. As  the  ball is  often invisible  while
possessed by the player, authors concentrate or court shot retrieval.

\subparagraph{Energy  function  minimization}  provide more  principled  way  of
finding a trajectory optimal with respect to a certain cost function.

\cite{Theobalt04} use very controlled environment  of darkened room to track the
baseball location and rotation by formulating an appropriate energy function.

\cite{Zhang08b} use the information about the nearest player, football
position, velocity, acceleration and trajectory length to train the
Adaboost classifier to filter a set of candidate trajectories, obtained by
applying the particle filter with the first-order linear motion to the detections.
The obtained trajectories were then taken into account when tracking people.
This approach uses people detection and tracking to generate ball trajectories
and obtain the final ball trajectory, but, unfortunately, no specific details of
ball tracking are given, and the evaluation is done only on short sequences.

\cite{Zhu07a} use the external information to detect the goal events
and their time stamp in football. Tracking is initialized using Viterbi algorithm
to link the detections and continues by using SVR particle filter until it is
lost. Based on ball and player locations and distances between each other,
attack is classified in one of six types.

\cite{Poiesi10} use  a detector-less approach  and generate a set  of basketball
candidates  by finding  the  points  of convergence  of  the  motion flow.  Such
candidates in each frame are joined into trajectories using a Kalman filter.

\cite{Wang14a} concentrate on a difficult part of tracking the ball while
it is in possession. Authors assume that ballistic parts of the ball
trajectories have already been extracted and concentrate on predicting which
person holds the ball based on the analysis of the game phase, player locations,
distances between them. Approach can be viewed as simultaneous ball tracking and
game phase estimation, but it is based on the correct extraction of ball
locations while it is not possessed, and the errors introduced at this stage can
not be recovered from. Authors show results on basketball and football datasets.

\cite{Wang14b} propose an integer programming formulation for tracking two types
of objects, one of which contains the  other, in a globally optimal way. Authors
show  their results  on several  datasets that  include basketball  and football
games. Since tracking  is done on a  discrete grid, precision of  3D tracking of
small  objects such  as  a ball  is limited.   Furthermore,  assumptions that  a
containee object  (ball) is  not visible  while it is  possessed by  a container
object (person) is clearly violated in the sports domain.

\subparagraph{Approaches without  the ball}  concentrate on  detecting
interactions without relying on localizing the ball.

\cite{Kim12} predict regions of interest by generating a stochastic vector field
of motion  tendencies, somewhat similar to~\cite{Poiesi10}.  While these regions
can often contain the ball, no information about the tracking accuracy is given.

In hockey, for the purposes of game elements recognition \cite{Li09d} and
role assignment \cite{Lan12}, hockey puck is either manually detected
or ignored.

In handball \cite{Direkoglu12}, ball location is not used as a feature for
activity recognition and authors mention that having it might allow to
recognize ``more complex activity classes''.

In American football for the problem of recognition of plays
\cite{Li09c}, ball location is not present in the formulation  and for
people locations ground truth data was used.

In baseball \cite{Gupta09}, action  elements recognition is based
on  the  roles  of  the  players. Role  assignment  was  done  without  tracking
the  ball  and  could  benefit from  it.  

\subparagraph{Our approach}

To         sum         up,         many        of         the         approaches
\cite{Chen09a,Chakraborty13,Ohno00,Leo08,Zhu07a}   use  domain   knowledge-based
heuristics  but  do  not  exploit  external information  such  as  the  location
of  the   players  or   the  state   of  the  ball.   In  many   sports  authors
\cite{Li09d,Direkoglu12,Lan12,Li09c,Gupta09}  concentrate on  features based  on
the motion flow, player locations and roles, etc. to understand the higher level
semantic of the game. Such applications would benefit greatly from the automatic
ball tracking that agrees with higher level semantics and is domain-independent.

Our work is most similar to the following works:
\begin{itemize}
  \item  We exploit the ability to express container-containee relations as
  an integer program, similarly to \cite{Wang14b}. However,
  our formulation of the tracking problem involves simultaneous tracking and
  state estimation, physical constraints and tracking in the continuous domain,
  that are not present in this work.
  \item We have the ball states similar to the ones in \cite{Ren08}, but
  we estimate them simultaneously with tracking and learn rather than define
  their features. Similarly to \cite{Zhang08b}, we weight the
  detections based on the features of location and distance to the nearest
  player and learn the likelihood of each individual detection, but our
  formulation includes the state of the ball.
  \item In a more general scheme of things, our approach learns the motion and
  appearance contexts of a ball and uses them to predict the location when it is
  unobserved, which is similar to what \cite{Ali07a} does in aerial
  videos. However, we don't operate under the assumption that the context can be
  unambiguously recovered from each frame. Ball states call also be viewed as
  learned \textit{supporters}, as defined by \cite{Grabner10}, but the
  our case of ball tracking we have full occlusions much more often and
  partial occlusions much more rarely.
\end{itemize}
}

\comment{
some  sports (\eg  tennis~\cite{Yan08} and  golf~\cite{Lepetit03a}, due  to rare
occlusions of the ball and its  discriminative appearance features), this is not
the  case  for many  team  sports.  Structured nature  of  data  in such  sports
gave  rise  to  many approaches  of  tracking  objects  with  the help  of  high
level  semantic cues.  Several works  enhance  tracking by  estimaing the  state
of  the  game~\cite{Liu13},  or  the  role of  the  player  and  the  formations
of  players~\cite{Lucey13}. For  the ball  tracking,~\cite{Wang14a} proposed  an
approach to track the ball while it  is in possession by using information about
the state of the game and players location. Our approach is similar to the above
in  the  sense  that  we  are  estimating the  state  of  the  ball  to  enhance
tracking  it. It  also  has  similarities to  more  general tracking  approaches
of~\cite{Ali07a}, as we  learn the motion and appearance context  of the ball to
predict  its location  while it  is unobserved.  However, we  don't assume  that
the  context can  be  unambiguously recovered  from each  frame.  Our states  of
the  ball  can  also  be  viewed  as  learned  \textit{supporters},  as  defined
by~\cite{Grabner10},  but for  the sports  scenario we  can not  claim that  the
relative target / features location is fixed over short time intervals. Below we
describe approaches more specific to ball tracking.
}


\section{Problem Formulation}
\label{sec:problem}

We  consider  scenarios   where  there  are  several   calibrated  cameras  with
overlapping fields  of view capturing  a substantial  portion of the  play area,
which means  that the  apparent size  of the  ball is  generally small.  In this
setting, trajectory growing methods do not  yield very good results both because
the  ball is  occluded too  often by  the players  to be  detected reliably  and
because its  being kicked or thrown  by them result in  abrupt and unpredictable
trajectory changes.

To remedy  this, we explicitly  model the interaction  between the ball  and the
players as well  as the physical constraints  the ball obeys when  far away from
the players. To this end, we first  formulate the ball tracking problem in terms
of a maximization  of a posteriori probability. We then  reformulate it in terms
of an  integer program. Finally,  by adding  various constraints, we  obtain the
final problem formulation that is a Mixed Integer Program.

\comment{We then show how this maximization can be reformulated as a constrained
network flow problem, which eventually results in a Mixed Integer Program (MIP).
\pfrmk{Should we mention here that we  track the players?} \amrmk{We can mention
here that  while tracking the  ball, we  also simultaneously track  the players,
based on their tracklets or ground truth trajectories?} \pfrmk{It's mentioned in
the  intro  and  the  caption  of  Fig. 1.  Whathever  you  say,  it  should  be
consistent.}}

\begin{figure}
\begin{center}
  \includegraphics[width=\columnwidth]{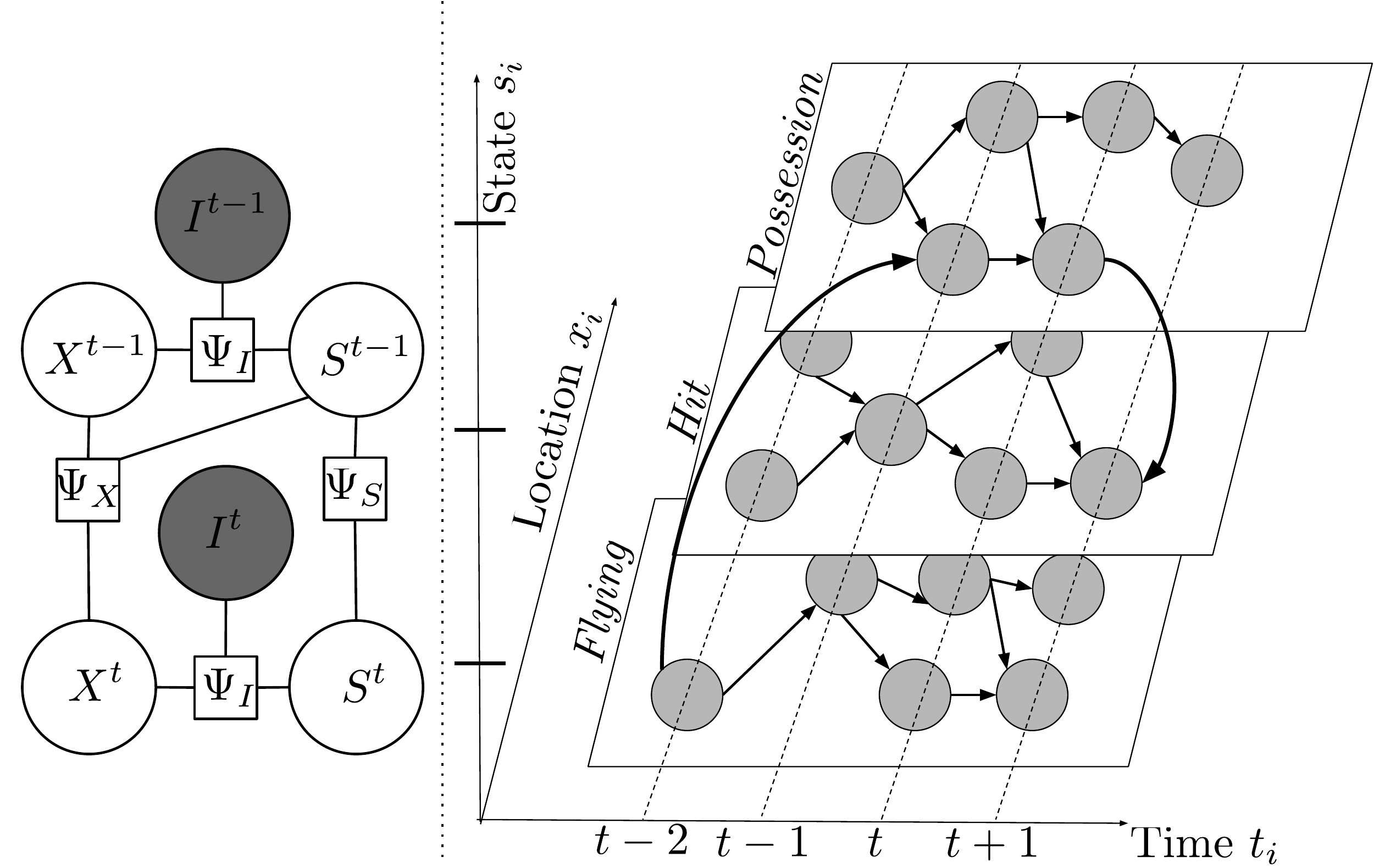}\\
  \hspace{-2cm} (a) \hspace{3.1cm} (b)
\end{center}
\vspace{-0.5cm}
\caption{Graphical models. \textbf{(a)} Factor graph for ball tracking.  At each
  time instant  $t$, we consider the  ball location $X^t$ and  state $S^t$ along
  with  the available  image evidence  $I^t$.  \textbf{(b)}  Ball graph  used to
  formulate the  integer program.  To  each node  $i$, is associated  a location
  $x_i$, a state $s_i$, and a  time instant $t_i$.  The relationship between the
  variables in both graphs is spelled out in Eqs~\ref{eq:ipEq}(d,e).}
\label{fig:factorGraph}
\end{figure}

\begin{table}[t]
\begin{small}
\centering 
\begin{tabular}{  r | l  } 
\hline
$T,K$ & Number of temporal frames and ball states \\
\hline
$I^t$ & Image evidence at time $t$\\
\hline
$X^t,S^t$ & Discrete location and state of the ball at time $t$ \\
\hline
$P^t$ & 3D coordinates of the ball at time $t$\\
\hline
$i,j,k,l$ & Node indices in the ball or players graph\\
\hline
$V_b,V_p$ & Sets of nodes in ball and player graphs\\
\hline
$E_b,E_p$ & Sets of edges in the  ball and  player graphs\\
\hline
$x_i,s_i,t_i$ & Discrete location, state, and time of node $i$\\
\hline
$S_b$ & Special node for the ball at $t=0$\\
\hline
$S_p,T_p$ & {\it Source} and {\it sink} nodes of player trajectories\\
\hline
$f_i^j,p_i^j$ & Number of balls and players moving from $i$ to $j$\\
\hline
$c_{bi}^j$,$c_{pi}^j$ & Ball and player transition costs from $i$ to $j$\\
\hline
\hspace{-1.5mm}$\Psi_X,\Psi_S,\Psi_I$ & Position, state, image evidence potentials\\
\hline
$\psi$ & Potential of local image evidence\\
\hline
$D_l$ & Max. permissible distance between $X^t$ and $P^t$\\ 
\hline
$D_p$ & Max. permissible distance for ball possession\\
\hline
$A^{s,c},B^{s,c}$ & Physics-based constants for state $s$, axis $c$\\
$C^{s,c},F^{c,s}$ \\
\hline
$O^{s,c}$ & Constraint-free locations for state $s$ and axis $c$\\ 
\hline
$\mathbb{F}$ & Permissible ball locations and state sequences\\
\hline
\end{tabular}
\caption{ Notations}
\vspace{-6mm}
\label{tab:notations} 
\end{small}
\end{table}

\subsection{Graphical Model for Ball Tracking}
\label{sec:graphModel}

We model the  ball tracking process from one  frame to the next in  terms of the
factor graph  depicted by  Fig.~\ref{fig:factorGraph}(a).  We associate  to each
instant  $t \in  \left\{1\ldots  T\right\}$ three  variables  $X^t$, $S^t$,  and
$I^t$, which respectively represent the 3D ball position, the state of the ball,
and the  available image evidence. When  the ball is within  the capture volume,
$X^t$  is a  3D vector  and $S^t$  can take  values such  as \textit{flying}  or
\textit{in\_possession},  which   are  common   to  all   sports,  as   well  as
sport-dependent ones,  such as  \textit{strike} for volleyball  or \textit{pass}
for basketball.   When the ball is  not present, we  take $X^t$ and $S^t$  to be
$\infty$ and \textit{not\_present} respectively. These  notations as well as all
the others we use in this paper are summarized in Table~\ref{tab:notations}.

Given the conditional  independence assumptions implied by the  structure of the
factor  graph of  Fig.~\ref{fig:factorGraph}(a), we  can formulate  our tracking
problem as one of maximizing the energy function%
\vspace{-3mm}
\begin{small}
\begin{eqnarray}
  \Psi(X,S,I) &  = &  {1\over Z} \Psi_I(X^1,S^1,I^1)  \prod\limits_{t=2}^T \Big[
    \Psi_X(X^{t-1},S^{t-1},X^t) \nonumber \\ 
& & \Psi_S(S^{t-1},S^t)\Psi_I(X^{t},S^{t},I^{t}) \Big] \label{eq:factors}
\end{eqnarray}
\end{small}%
expressed in terms  of products of the following  potential functions:\comment{ over the
maximal            cliques           of            the           \am{appropriate
  MRF}\comment{graph}~\cite{Bishop06}.  \pfrmk{In~\cite{Bishop06}, they  are the
  cliques of the factor graph. Why did you change it?}}
\begin{itemize}

    \item $\Psi_I(X^{t},S^{t},I^{t})$  encodes the correlation between  the ball
        position, ball state,  and  the \comment{observed}  image  evidence. 

    \item $\Psi_S(S^{t-1},S^t)$ models the  temporal smoothness of states across
adjacent  frames. 

  \item  $\Psi_X(X^{t-1}, S^{t-1},  X^t)$  encodes the  correlation between  the
  state of the ball  and the change of ball position from one frame to the next
  one.

  \item $\Psi_X(X^1,S^1,X^2)$ and $\Psi_S(S^1,S^2)$  include priors on the state
  and position of the ball in the first frame.

\end{itemize}%
In  practice, as will  be discussed in Sec.~\ref{sec:learning},  the $\Psi$
functions are learned from training data.
Let $\mathbb{F}$  be the  set of  all possible sequences  of ball  positions and
states.   We consider  the log  of  Eq.~\ref{eq:factors} and  drop the  constant
normalization factor $\log Z$. We, therefore, look for the most likely sequence of
ball positions and states as%

\vspace{-5mm}
\begin{small}
\begin{align}
\label{eq:mainEq}
& (X^*, S^*) = \arg\max \limits_{(X,S) \in \mathbb{F}} \sum \limits_{t=2}^{T} \Big[ \log \Psi_X(X^{t-1},S^{t-1},S^t) + \\ \nonumber
& \log  \Psi_S(S^{t-1},S^t)  +  \log  \Psi_I(X^{t},S^{t},I^{t})  \Big]  +  \log
\Psi_I(X^1,S^1,I^1) \; . \\ \nonumber
\end{align} 
\end{small}%
\vspace{-1cm}

In the  following subsections,  we first  reformulate this  maximization problem
as  an  integer   program  and  then  introduce   additional  physics-based  and
\textit{in\_possession} constraints.

\subsection{Integer Program Formulation}
\label{sec:ip}

To  convert the  maximization  problem of  Eq.~\ref{eq:mainEq}  into an  Integer
Program  (IP), we  introduce the  {\it ball  graph} $G_b=(V_b,E_b)$  depicted by
Fig.~\ref{fig:factorGraph}(b). $V_b$  represents its nodes, whose  elements each
correspond to a location $x_i \in \mathbb{R}^3$, state $s_i \in \{1,\cdots,K\}$,
and time index $t_i \in \{1,\cdots,T\}$. {In practice, we instantiate as many
as there are possible states at every time step for every actual and potentially
missed ball detection.  Our approach to hypothesizing such  missed detections is
described in  Sec.~\ref{sec:Graphs}.} $V_b$ also contains  an additional
node $S_b$ denoting  the ball location before the first  frame. $E_b$ represents
the edges of $G_b$ and comprises all pairs of nodes corresponding to consecutive
time instants and whose locations are  sufficiently close for a transition to be
possible.

Let  $f_i^j$ denote the number of  balls moving from $i$ to $j$
and $c_{bi}^j$ denote the corresponding cost.
The maximization  problem of Eq.~\ref{eq:mainEq}  can be
rewritten as%

\vspace{-1.5mm}
\begin{small}
  \begin{equation}
   \text{maximize} \displaystyle\sum\limits_{(i,j) \in E_b}f_i^jc_{bi}^j \; , \label{eq:ipEq}
   \end{equation}
   \vspace{-2mm}
   \text{where} \\[3mm]
    $c_{bi}^j = \log  \Psi_X(x_i,s_i,x_j) + \log \Psi_S(s_i,s_j) + \log\Psi_I(x_j,s_j,I^{t_j}) ,$\\[2mm]
   \text{subject to} \\
   \vspace{-3mm}
   \begin{equation*}
\begin{array}{ll@{}ll}
            &\textit{(a)} &\mbox{\hspace{3mm}}f_i^j \in \{0,1\} &\mbox{\hspace{-0mm}}\forall (i,j) \in E_b\\
            &\textit{(b)} &\mbox{\hspace{3mm}}\sum\limits_{(i,j) \in E_b, t_j=1}f_i^j = 1                      &                               \\
            &\textit{(c)} &\mbox{\hspace{3mm}}\sum\limits_{(i,j) \in E_b}f_i^j=\sum\limits_{(j,k) \in E_b}f_j^k  &\mbox{\hspace{-0mm}}\forall j \in V_b: 0 < t_j < T      \\
            &\textit{(d)} &\mbox{\hspace{3mm}}X^t = \sum\limits_{(i,j) \in E_b,t_j=t}f_i^jx_j                  &\mbox{\hspace{-0mm}}\forall t \in 1,\cdots,T       \\
            &\textit{(e)} &\mbox{\hspace{3mm}}S^t = \sum\limits_{(i,j) \in E_b,t_j=t}f_i^js_j                  &\mbox{\hspace{-0mm}}\forall t \in 1,\cdots,T       \\   
            &\textit{(f)} &\mbox{\hspace{3mm}}(X,S) \in \mathbb{F} & \\
\end{array}
\end{equation*}
\end{small}%
We optimize  with respect to  the $f_i^j$, which  can be considered  as flow
variables.  The  constraints of  Eqs.\ref{eq:ipEq}(a-c) ensure that  at every
time frame there exists  only one position and one state to  which the only ball
transitions from the  previous frame.  The constraint  of Eq.\ref{eq:ipEq}(f) is
intended  to  only  allow  feasible  combinations of  locations  and  states  as
described by the set $\mathbb{F}$, which we define below.

\subsection{Mixed Integer Program Formulation}
\label{eq:MIP}

Some ball states impose first and  second order constraints on ball motion, such
as zero  acceleration for the freely  flying ball or zero  vertical velocity and
limited negative acceleration for the rolling ball.  Possession implies that the
ball must be near the player.

In this section,  we assume that the players' trajectories  are available in the
form  of a  {\it player  graph}  $G_p=(V_p,E_p)$ similar  to the  ball graph  of
Sec.~\ref{sec:ip}  and whose  nodes comprise  locations $x_i$  and time  indices
$t_i$. In practice, we compute it  using publicly available code as described in
Sec.~\ref{sec:playerGraph}.

Given  $G_p$, the  physics-based and  possession constraints  can be  imposed by
introducing auxiliary continuous  variables and expanding constraint   of Eq.~\ref{eq:ipEq}(f), as follows.

\vspace{-3mm}\paragraph{Continuous Variables.}

The $x_i$ represent  specific 3D locations where the ball  could potentially be,
that is, either actual ball detections or hypothesized ones as will be discussed
in  Sec.~\ref{sec:ballGraph}. Since  they cannot  be expected  to be  totally
accurate, let the continuous variables $P^t=(P_x^t,P_y^t,P_z^t)$ denote the true
ball position of at time $t$. We impose%

\vspace{-4mm}
\begin{small}
\begin{equation}  
    \label{eq:discreteCont} 
    ||P^t  - X^t|| \le D_l 
\end{equation}
\end{small}%
where $D_l$ is  a constant that depends  on the expected accuracy  of the $x_i$.
These continuous variables can then be used to impose ballistic constraints when
the ball is in flight or rolling on the ground as follows.


\vspace{-3mm}\paragraph{Second-Order Constraints.}

For each state $s$ and coordinate $c$ of $P$, we can formulate a
second-order constraint of the form%

\vspace{-4mm}
\begin{small}
\begin{align}
\label{eq:secondOrder}
& A^{s,c} (P^t_c - 2 P^{t-1}_c + P^{t-2}_c) + B^{s,c} (P^t_c - P^{t-1}_c) + \\ 
& C^{s,c} P^t_c - F^{s,c} \le K  (3 - M^t_{s,c} - M^{t-1}_{s,c} - M^{t-2}_{s,c})
\; ,
\nonumber \\
& \mbox{where} \, \, M^t_{s,c} =  \sum\limits_{(i,j)\in E_b, t_j=t,s_j=s, x_j \not \in
    O^{s,c}}f_i^j \; , \nonumber
\end{align}
\end{small}%
$K$ is a large positive constant and $O^{s,c}$ denotes the locations
where  there  are scene elements with which the ball  can collide, 
such as those near the basketball hoops or close to  the  ground.
Given the constraints of
Eq.~\ref{eq:ipEq}, $M^t_{s,c}$,  $M^{t-1}_{s,c}$, and  $ M^{t-2}_{s,c}$  must be
zero or one. This implies that right side of the above inequality is either zero
if  $M^t_{s,c}  =  M^{t-1}_{s,c}  =  M^{t-2}_{s,c}   =  1$  or  a  large  number
otherwise. In  other words,  the constraint  is only  effectively active  in the
first case, that is,  when the ball consistently is in a  given state. When this
is the case, $(A^{s,c},B^{s,c},C^{s,c}$,$F^{s,c})$ model the corresponding
physics. For  example, when  the ball  is in the  \textit{flying} state,  we use
$(1,0,0,{-g \over fps^2})$ for the $z$  coordinate to model the parabolic motion
of an object subject to the sole force of gravity whose intensity is $g$. In the
\textit{rolling} state, we use $(1,0,0,0)$ for  both the $x$ and $y$ coordinates
to denote a  constant speed motion in  the $xy$ plane. In both cases, we neglect
the effect of friction.  We give more details for all states we represent in the
supplementary materials. Note that we turn  off these  constraints altogether
at locations in $O^{s,c}$.

\comment{
We turn  off these  constraints altogether  at locations  where there  are scene
elements which the ball  can collide, such as very near  the basketball hoops or
very  close to  the  ground.  \am{Such locations  are  denoted  by $O^{s,c}$  in
Eq.~\ref{eq:secondOrder}.}
}

\comment{
Finally, $M^t_{s,x}\equiv \mathbb{1}(S^t=s \land X^t \not \in O^s_x)$. If $O^s_x
=\emptyset$, then  $M^t_{s,x}\equiv \mathbb{1}(S^t=s)$, which means  that we are
placing the constraint  iff. for the last  3 consecutive frames the  ball was in
the state $s$. However,  there are particular locations in which  we do not want
to place the constraint on the ball. They correspond to the known scene elements
with which  the ball  can collide. For  example, for basketball  we do  not want
to  impose  constraints  when  the  ball  is in  the  vicinity  of  the  basket.
Furtherwore, we  do not  want to  impose constraints  on the  $z$ axis  when the
ball  is flying  and  is near  the  ground  (height less  that  $H$, this  means
the  ball  will  bounce  off  the  ground  and  continue  to  fly).  We  express
that by  saying that  set of  model-breaking locations  $O^{\textit{flying}}_z =
\{x_i=(x_{i,1},x_{i,2},x_{i,3})^\intercal  \in V_b:  x_{i,3} <  H\}$. Also  note
that since such sets  are defined separately for each axis,  we will continue to
impose zero acceleration in horizontal plane  while the ball is flying. Detailed
definition  of $O^s_x$  for  all  states and  axes  are  given in  supplementary
materials. \am{There  could be multiple  constraints of  such form for  a single
state, for example to limit the acceleration both from below and from above.}
}

\comment{ For each state $s$ and axis  $x$ we express second order constaints on
the motion  of the ball  in this state  on this axis  as a linear  constraint as
follows: $M >>  0$ is a number far  greater than any value in  the expression on
the left hand side of the inequality and $M^t_{s,x}$ is the state of the ball at
time $t$. $A^{s,x},  B^{s,x}, C^{s,x}, F^{s,x}$ are constants  that describe the
trajectory  of  the  ball.  Therefore,  contraint  will  only  be  enforced  iff
for  3 consecutive  frames  the ball  was  in the  state  $s$, corresponding  to
$M^t_{s,x}=M^{t-1}_{s,x}=M^{t-2}_{s,x}=1$.  \subparagraph{Collision  with  scene
elements}  To account  for scenarios  when  the ball  does not  follow a  single
physically viable model, but collides with the element of the scene, such as the
basket, net or the floor, we modify the definition of $M^t_{s,x}$ to include the
flow of  the ball not  through all nodes  with state $s$  at time $t$,  but only
through the nodes that do not lie near the elements of the scene. More formally,
if $O_x$ is the set of nodes that correspond to ball locations near the elements
of the scene that allow the ball to change the trajectory in axis $x$, then
\begin{equation}
  \label{eq:sceneElements}
  M^t_{s,x} = \sum\limits_{(i,j)\in E, t(j)=t,s(j)=s, j \not \in O_x}f_i^j \; .
\end{equation}
By defining  the set of  scene element  nodes with respect  to the axis,  we can
allow the change in  velocity only in a single axis, such as  when the ball hits
the floor, the motion should stay linear, but the Z velocity may change. }

\vspace{-3mm}\paragraph{Possession constraints.}

While the ball is in possession of  a player, we do not impose any physics-based
constraints. Instead, we  require the presence of someone  nearby. The algorithm
we  use for  tracking the  players~\cite{Berclaz11} is  implemented in  terms of
people flows that we denote as $p_i^j$ on a player graph $G_p=(V_p,E_p)$
that plays the  same role as the ball  graph. The $p_i^j$ are taken  to be those
that
\vspace{-3mm}
\begin{small}
  \begin{equation}
    \text{maximize} \displaystyle\sum\limits_{(i,j) \in E_p}p_i^jc_{pi}^j \;
    , \label{eq:peopleEq}
  \end{equation}
\text{where} \hspace{1.2mm} $c_{pi}^j =  {\log P_p(x_i|I^{t_i})\over 1-\log P_p(x_i|I^{t_i})} \; ,$\\
\text{subject to} \\
  \begin{equation*}
  \vspace{-1mm}
\begin{array}{ll@{}ll}
            
&\textit{(a)} &\mbox{\hspace{3mm}}p_i^j \in \{0,1\} &\forall (i,j) \in E_p\\
&\textit{(b)} &\mbox{\hspace{3mm}}\sum\limits_{i:(i,j) \in E_p}p_i^j \le 1                              &\forall j \in V_p\setminus\{S_p\}           \\
&\textit{(c)} &\mbox{\hspace{3mm}}\sum\limits_{(i,j) \in E_p}p_i^j=\sum\limits_{(j,k) \in E_p}p_j^k     &\forall j \in V_p\setminus\{S_p,T_p\} \ .       \\ 
\end{array}
\end{equation*}
\end{small}%
Here $P_p(x_i|I^{t_i})$  represents the output of  probabilistic people detector
at location  $x_i$ given  image evidence  $I^{t_i}$. $S_p,T_p  \in V_p$  are the
source and  sink nodes that  serve as starting  and finishing points  for people
trajectories,  as  in~\cite{Berclaz11}.  In  practice we  use  the  publicly
  available code of~\cite{Fleuret08a} to compute the probabilities $P_p$ in each
  grid cell of discretized version of the court.

Given the ball flow variables $f_i^j$ and people flow ones $p_i^j$, we express the
\textit{in\_possession} constraints as
\begin{small}
\begin{equation}
\sum\limits_{\substack {(k,l) \in E_p,t_l=t_j, \\ ||x_j-x_l||_2 \le D_p}}\hspace{-1mm}p_k^l  \ge \sum\limits_{i:(i,j) \in E_b}\hspace{-1mm}f_i^j   \hspace{4.2mm} \forall j:s_j\equiv{\scalebox{0.7}{in\_possession}}\; ,
\label{eq:possConst}
\end{equation}
\end{small}%
where $D_p$  is the maximum possible distance between the player and the ball location 
when the player is in control of it, which is sport-specific. 

\comment{
physical  constraints  on the  ball  trajectory  have  to be  imposed.  However,
possession requires  a person present near  the position of the  ball. We assume
that information  about players is  given in the form  of the graph  of possible
player  locations (as  before,  each node  in  the graph  is  identified by  the
location $x_i$  and time index  $t_i$. $s_i$ are  not defined for  people graph)
along with  allowed transitions between those  states and their cost.  Note that
this definition  includes a  set of  detections (with  allowed transitions  in a
certain spacial  vicinity in  adjacent frames,  \eg~\cite{Berclaz11}), a  set of
tracklets   (with   a   set   of  additional   linking   paths   between   them,
\eg~\cite{Wang14b}), or simply  a set of true player trajectories.   We denote a
set of  all nodes as  $V_p$, and all possible  transitions as $E_p$.  $V_p$ also
includes two additional nodes $S_p$ and  $T_p$.  $E_p$ includes edges from $S_p$
to all  locations where player trajectories  can possibly begin, and  edges from
all locations where player trajectories could possibly end to $T_p$. $f_i^j$ are
binary variables defined  for all $(i,j) \in E_p$, and  represent the transition
of a  player from location  $i$ to location $j$.   $c_i^j$ is a  cost associated
with the transition.
}

\vspace{-3mm}\paragraph{Resulting MIP.}

Using    the    physics-based     constraints    of    Eq.~\ref{eq:discreteCont}
and~\ref{eq:secondOrder}     and      the     possession      constraints     of
Eq.~\ref{eq:possConst}  along  with  the  formulation of  people  tracking  from
Eq.~\ref{eq:peopleEq}  to   represent  the  feasible   set  of  states   $F$  of
Eq.~\ref{eq:ipEq}(f) yields the MIP
\vspace{-1mm}
\begin{small}
  \begin{equation}
\begin{array}{ll@{}ll}
\mbox{\hspace{-5mm}}&\text{maximize}\displaystyle\sum\limits_{(i,j) \in E_b} f_i^jc_{bi}^j + 
                             \sum\limits_{(i,j)        \in       E_p} p_i^jc_{pi}^j \\ 
\mbox{\hspace{-5mm}}&\text{subject            to            the            constraints            of
    Eqs.\ref{eq:ipEq}(a-e),~\ref{eq:discreteCont},~\ref{eq:secondOrder},~\ref{eq:peopleEq}(a-c), and~\ref{eq:possConst}.} 
\end{array}
\label{eq:finalEq} 
\end{equation}
\end{small}
In  practice,  we  use  the   Gurobi~\cite{Gurobi}  solver  to  perform  the
  optimization. Note that we can either  consider the people flows as given and
optimize only  on the ball flows  or optimize on both  simultaneously.  {We will
  show in  the results section that  the latter is only  slightly more expensive
  but    yields    improvements    in    cases    such    as    the    one    of
  Fig.~\ref{fig:motivation}.}

\comment{
\begin{small}
\begin{equation}
\begin{array}{ll@{}ll}
&\text{maximize}\displaystyle\sum\limits_{(i,j) \in E_b \cup E_p}f_i^jc_i^j                                        \\                              
&\text{subject to}                                                                                                   \\ 
&\text{(\ref{eq:ipEq}b-f,~\ref{eq:discreteCont},~\ref{eq:secondOrder})} &                                            \\                               
&f_i^j \in \{0,1\}                                                     &\forall (i,j) \in E_p                       \\
&\sum\limits_{i:(i,j) \in E_p}f_i^j \le 1                              &\forall j \in V_p\setminus\{S_p\}           \\
&\sum\limits_{(i,j) \in E_p}f_i^j=\sum\limits_{(j,k) \in E_p}f_j^k    &\forall j \in V_p\setminus\{S_p,T_p\}       \\ 
&\sum\limits_{\substack {(k,l) \in E_p,t_l=t_j, \\ ||x_j-x_l||_2 \le D_p}}\hspace{-1mm}f_k^l  \ge \sum\limits_{i:(i,j) \in E_b}\hspace{-1mm}f_i^j   \hspace{4.2mm}                         
&\forall j:s_j\equiv\small{\scalebox{0.7}{\texit{in\_possession}}}      \\
\end{array}
\label{eq:finalEq}
\end{equation}
\end{small}%
where $D_p$  in  the last  constraint  indicate  maximum  distance  of ball,  that  is
possessed by the player, to the player. It is different for different sports and
is learned from the training data.  Last constraint allows ball to transition to
a  position in  a vicinity  of position  where the  player has  moved to.  Three
previous constraints force  equal number of people to enter  and leave location,
and limit this  number to 1. Note that  in the case if true  trajectories of the
players are given, we set $c_i^j$  in players graph to same (arbitrary) positive
value, and  in the optimal  solution $f_i^j=1 \forall  (i,j) \in E_p$.  In other
case  we define  of  the  edge to  be  $c_i^j=\log({P_p(x_j|I^{t_j})  \over 1  -
P_p(x_j|I^{t_j})})$, where  $P_p(x|I)$ is the  output of the people  detector in
location $x$. More details can be found in the original work of~\cite{Berclaz11}
describing this approach.  Eq.~\ref{eq:finalEq} is the final  formulation of our
tracking and state estimation problem.
}

\comment{
\subparagraph{Simultaneous tracking  of the ball  and the players} Note  that we
have defined $E_p$ as a graph of all transitions of players in the ground truth,
and $f_p$  are constants representing  those transitions. In the  fasion similar
to~\cite{Wang14b},  we can  also define  $V_p$  to be  the set  of all  possible
locations of the players,  $E_p$ to be the set of  all possible transitions, and
$f_p$  to  be  the  binary  variables representing  the  transitions.  To  those
variables, we  apply constraints  similar to  (\ref{eq:ipEq}b,d) to  ensure that
number  of  people  that  enter  and exit  the  location  is  equal.  Constraint
(\ref{eq:ipEq}c) for people takes  the following form: $\sum\limits_{i:(i,j) \in
E_p}f_i^j  \le 1,\forall  j \in  V_p\setminus\{S_p\}$, ensuring  that only  one
person can occupy each certain location, but not limiting total number of people
in the  scene. For locations $i,j$  in graph of people  locations $c_i^j$ would
then  represent the  cost  associated  with the  person  being  present at  $j$,
and  the objective  function  would  take the  form  of $\sum\limits_{(i,j)  \in
E_b}f_i^jc_i^j+\sum\limits_{(i,j) \in E_p}f_p(i,j)c(i,j)$.
}

\comment{
We consider  two  possible  scenarios here.  In  one,
ground truth  positions of  the players  are given.  In the  second, we  use the
approach of~\cite{Wang14b},  where people  detections are  joined using  the KSP
algorithm~\cite{Berclaz11} into tracklets, which are connected using the Viterbi
algorithm to create  a graph of possible  ball locations. In both  cases, we can
treat all possible players locations as the nodes in the graph $P(V_p,E_p)$ with
flows $f_p(i,j)$  representing the trajectories  of the players. To  this graph,
same  constraints of  flow  conservation  and unit  flow  apply. The  possession
constraint can therefore be expressed as follows:

\begin{equation}
\label{eq:possessionConstr}
|L^t - \sum\limits_{(i,j) \in E_p}x(j)f_p(i,j)| \le D_p
\end{equation}

$D_p$ is the maximum distance from the  ball to the ground plane location of the
player possessing  it and  it is learned  from the ground  truth data.  The edge
costs   are  computed   as  $c_i^j   =  log   {P_p(x(j)|I^{t(j)})  \over   1  -
  P_p(x(j)|I^{t(j)})}$ for edges between the frames, with $P_p$ being the output
of the people  detector. For edges from  the sink and to the  source, a constant
penalty associated with the prior probability  of person entering or exiting the
scene is used as a weight. More  details can be found in~\cite{Wang14b}. In case
of a given ground truth, we can  assume all edges to have any arbitrary positive
weight.

\subsection{Final formulation}

Adding                                                               constraints
(\ref{eq:discreteCont},~\ref{eq:secondOrder},~\ref{eq:sceneElements},~\ref{eq:possessionConstr})
finishes  the   formulation  of  the   integer  linear  program   equivalent  to
(\ref{eq:mainEq}-\ref{eq:locationConstr}):

\begin{align}
\text{maximize} \sum\limits_{(i,j)\in \{E,E_p\}}f_i^jc_i^j
\label{eq:mainFlow} \\
\text{\textit{subject to}
(\ref{eq:conservationConstr},\ref{eq:unitFlow},\ref{eq:discreteCont},\ref{eq:secondOrder},\ref{eq:sceneElements},\ref{eq:possessionConstr})}
\end{align}

Deterministic relationship between $X^t$ and the flow is given by $X^t=P^t$, and
between $S^t$ and the flow is given by $S^t=s:M^t_s=1$.
}


\section{Learning the Potentials}
\label{sec:learning}

In this section, we define  the potentials introduced in Eq.~\ref{eq:mainEq} and
discuss how their  parameters are learned from training data.  They are computed
on  the nodes  of the  ball graph  $G_b$ and  are used  to compute  the cost  of
the  edges,  according to  Eq.~\ref{eq:ipEq}.  We  discuss its  construction  in
Sec.~\ref{sec:ballGraph}.

\vspace{-3mm}
\paragraph{Image evidence potential $\Psi_I$.}

It models  the agreement between  location, state,  and the
image evidence. We write
\begin{small}
\begin{eqnarray}
\Psi_I(x_i,s_i,I) & = & \psi(x_i,s_i,I)\prod_{\mathclap{\substack{j \in V_b:t_j=t,\\ (x_j,s_j)\not=(x_i,s_i)}}}\;\Big(1-\psi(x_j,s_j,I)\Big)\;,\nonumber \\
\psi(x,s,I) & = & \sigma_s(P_b(x|I)P_c(s|x,I))\;, \label{eq:factorZ}\\
\sigma_s(y) & = & {1\over 1+e^{-\theta_{s0}-\theta_{s1}y}}\;, \nonumber 
\end{eqnarray}
\end{small}%
where  $P_b(x)$ represents  the  output of  a ball  detector  for location  $x$,
$P_c(s|x,I)$ the  output of  multiclass classifier that  predicts the  state $s$
given the  position and the local  image evidence.  $psi(x,s,I)$ is  close to
one  when the  ball is  likely to  be located  at $x$  in state  $s$ with  great
certainty  based  on  image  evidence  only  and  its  value  decreases  as  the
uncertainty of either estimates increases.

In  practice,  we  train  a  Random  Forest  classifier~\cite{Breiman01}  to
estimate $P_c(s|x,I)$.  As features, it  uses the  3D location of  the ball.
Additionally, when  the player trajectories  are given,  it uses the  number of
people in its vicinity as a feature. When simultaneously tracking the players and
the ball,  we instead use the integrated outputs of the people  detector in the vicinity  of the
ball. We give additional details in the supplementary materials.

  The parameters  $\theta_{s0},\theta_{s1}$ of the logistic  function $\sigma_s$
are  learned from  training data  for each  state $s$.  Given the  specific ball
detector  we  rely  on,  we  use  true and  false  detections  in  the  training
data  as  positive and  negative  examples  to  perform a  logistic  regression.

\vspace{-3mm}
\paragraph{State  transition potential $\Psi_S$.}  
We define it  as the  transition probability between states, which we learn
from the training data, that is:

\vspace{-4mm}
\begin{small}
\begin{equation}
\label{eq:factorS}
\Psi_S(s_i,s_j)=P(S^{t}=s_i|S^{t-1}=s_j)\;.
\end{equation}
\end{small}%
As noted in  Sec.~\ref{sec:graphModel}, potential for the  first time frame
has a special  form $P(S^2=s_i|S^1=s_j)P(S^1=s_j)$, where  $P(S^1=s_j)$ is the
probability of the  ball being in state  $s_j$ at arbitrary time  instant; it is
learned from the training data.

\comment{We  do  not use  any  prior  on the  probability  to  ensure that  some
transitions  are forbidden  (\eg probability  of transitioning  from rolling  to
flying is 0, as it would require ball to be possessed by the player in between).
For $Psi_S(S^1)$  we use the  prior probability  $P(S^t)$ also learned  from the
training data.}

\vspace{-3mm}
\paragraph{Location change potential  $\Psi_X$.} It models the transition of
the ball between two time instants. 
Let $D^s$ denote the maximum speed of the ball when in state $s$.
We write it as

\vspace{-3mm}
\begin{small}
\begin{equation}
\label{eq:factorX}
\Psi_X(x_i,s_i,x_j)=\mathbb{1}(||x_i-x_||_2 \le D^{s_i})\;.
\end{equation}
\end{small}%
For the \textit{not\_present} state,  we only allow transitions between the  
node representing the absent ball and the nodes near the border of the tracking area.
For the first frame the potential has an additional  factor of
$P(X^1=x_i)$,  ball location  prior, which  we assume  to be  uniform inside  of the
tracking area.

\comment{
$P_c(x|s,I)$  uses as the features  the 3D location
of the  ball and the number  of people in  certain vicinities (\eg 50cm,  2m for
basketball). It is learned using Random Forest~\cite{Breiman01}. We make use of
the symmetry  of playing  field to  reduce overfitting. We  use one  variable to
learn each  tree, due to small  amount of training data  and because intuitively
one feature  is often  enough to classify  the ball state  (\eg height  above 5m
$\to$ \textit{flying}; no people around $\to$ not \textit{in\_possession}). 
}

\comment{We learn the state classifier using random forest~\cite{Breiman01}. As
features, we use the location of the ball $(x,y,z)$, and the number of people in
certain vicinities (\eg 50cm, 2m for basketball). To reduce overfitting, we make
use of the fact that for sports games the field is symmetric with respect to the
rotation of  180 degrees over the  center of the  field. We use one  variable to
learn  each tree,  due  to small  amount  of  training data  and  the fact  that
intuitively  one can  often classify  the ball  state ball  based on  the single
feature (\eg if the  ball is high in the air, it is  likely flying; if there are
no people in the  vicinity of the ball, it is likely  not possessed, etc.). When
applying the classifier, if we track the ball and the players simultaneously, we
don't know the  actual locations of the  people. In that case  we substitute the
number  of people  by  the output  of  the people  detector,  integrated in  the
vicinity of  the ball location.  \am{Example of  a learned classifier  output is
presented in supplementary materials.}}

\comment{\begin{figure}
    \includegraphics[width=\columnwidth]{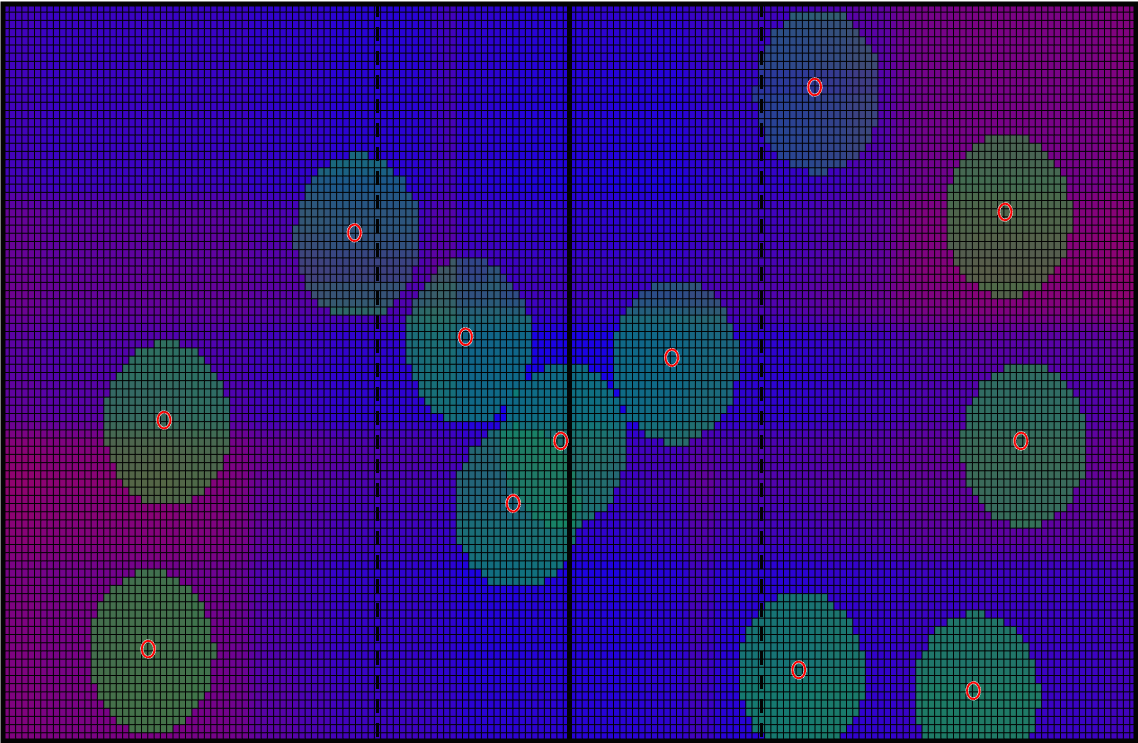}

    \caption{Example  of ball  classifier  output. People  are located  simiraly
    to  Fig.~\ref{fig:motivation}. Input  to  classifier is  $(x,y,z,p)$, $x,y$  are
    locations in  the field,  $z$ is  at constant  height of  2.5 meters,  $p$ is
    computed by integrating  number of people in the vicinity  of 1 meter. R,G,B
    elements of  color indicate probability  of being in  states \textit{flying},
    \textit{in\_possession},  \textit{hit}.  Area  around  people  is  attributed
    mostly to possession, with higher probability in the vicinity of two people.
    Area in the  middle of the court is predicted  mostly as \textit{hit}, while
    for areas from which players serve the ball it is mostly \textit{flying}. Best viewed in color.}

    \label{fig:classifierExample}
\end{figure}

}

\comment{An   example   output    of   such   classifier   can    be   seen   in
Fig.~\ref{fig:classifierExample}.}

\comment{
For each state  we train a logistic regression on  detections from training data
with a  single input $P_b(x|I)P_c(s|x,I)$,  and use the  output as the  value of
$\psi(x,s,I)$. $P_c(s|x,I)$ is  the output of ball state  classifier we describe
later. Such function has two desired properties.  First, it is higher when it is
likely that the  ball is indeed located at  $x$, and when it is  likely that the
ball is indeed  in state $s$. Second, learned regression  parameters serve as an
adjustment for different visibility of the ball and confidence in detections for
different states.  This reflects our  uncertainty about  the output of  the ball
detector: we  expect for  the state \textit{in\_possession}  to have  many false
detections on moving  people around the ball, as well  as many missed detections
due  to  occlusions,  compared  to \textit{flying};  the  cost  function  should
produce  values  of smaller  magnitude  for  the  former state.  Such  behaviour
of  ball detectors  is  not  specific to  ours  -  \eg~\cite{Zhang08b}. For  the
\textit{not\_present} state, $\psi(x,s=\emptyset,I)=\mathbb{1}(x=\infty)$.
}
\comment{
We wantψ(x,s,I) to be  higher when it is likely that the  ball is indeed located
at x, and when it is likely that the ball is indeed in state s. Additionally, we
want ψ  to reflect our  uncertainty when in particular  state the output  of the
ball detector is less informative (e.g.  we are more confident in the detections
when  the ball  is flying,  and want  the function  to be  higher when  the ball
detector output is high, compared to the same output of the ball detector in the
vicinity  of  the   person,  as  movement  of  people   often  produce  spurious
detections. Similarly, when  want to penalize ‘invisible’ flying  ball more than
an  ‘invisible’ ball  in possession,  as  in the  latter  case it  can often  be
occluded). Function ψ(x,  s, I) = σs(Pb(x|I)Pc(s|x, I)) possesses  both of those
properties (σ(x) =  1 ). P denotes the  output of 1+e−x c the  classifier of the
ball states.  The  parameters of a sigmoid serve as  an adjustment for different
visibility of  the ball  and confidence  of detections  in different  states. To
learn the parameters of the sigmoid for each  of the states, we use all true and
false  detections of  the ball  in the  training data  as positive  and negative
examples. For the state of  ball being absent, we define ψ(x, s =  ∅, I) = 1(x =
∞). Note that training  data is based on a specific  ball detector and therefore
learning  has  to  be  done  with  the  same  ball  detector  as  used  for  the
application. However,  it is not  specific to our  particular detector -  it has
been observed that other detectors also show a different distribution of outputs
depending on the ball state (e.g. [35]).}

\comment{  when in  particular state  the output  of the  ball detector  is less
informative  (\eg we  are more  confident  in the  detections when  the ball  is
flying, and  want the  function to be  higher when the  ball detector  output is
high, compared to  the same output of  the ball detector in the  vicinity of the
person, as movement of people often produce spurious detections. Similarly, when
want  to penalize  `invisible'  flying ball  more than  an  `invisible' ball  in
possession, as in  the latter case it can often  be occluded). Logistic function
$\psi(x,s,I)={1\over  1  + e^{-\theta_{s0}  -  \theta_{s1}P_b(x|I)P_c(s|x,I)}}$
possesses  the first  property,  and its  parameters $\theta_{s0},  \theta_{s1}$
serve as  an adjustment for different  visibility of the ball  and confidence of
detections  in different  states.  To learn  them  for each  of  the states,  we
use  all  true  and false  detections  of  the  ball  in the  training  data  as
positive and  negative examples. For the  state of ball being  absent, we define
$\psi(x,s=\emptyset,I)=\mathbb{1}(x=\infty)$. Note  that training data  is based
on a specific ball detector and therefore  learning has to be done with the same
ball  detector  as  used  for  the application.  However,  it  is  not  specific
to  our  particular  detector  -  it has  been  observed  that  other  detectors
also  show a  different  distribution of  outputs depending  on  the ball  state
(\eg~\cite{Zhang08b}). }

\comment{and detectors and  classifiers and describe the process  of learning it
  from the training data. After that we describe the detectors and classifier we
  use as part  of the potentials.  Training data contains  information about the
  state of the ball  and its location at a particular time  instance, as well as
  the locations of the players.}

\section{Building the Graphs}
\label{sec:Graphs}

Recall   from  Sections~\ref{sec:ip}   and~\ref{eq:MIP},  that   our
  algorithm operates on a ball and player graph. We build them as follows.

\subsection{Player Graph}
\label{sec:playerGraph}

To   detect   the   players,   we  first   compute   a   Probability   Occupancy
Map   on   a  discretized   version   of   the   court   or  field   using   the
algorithm  of~\cite{Fleuret08a}.  We  then  follow  the  promising  approach
of~\cite{Wang14b}. We use the K-Shortest-Path (KSP)~\cite{Berclaz11} algorithm
to  produce  tracklets,  which  are  short  trajectories  with  high  confidence
detections. To hypothesize  the missed detections, we use  the Viterbi algorithm
on the discretized grid to connect  the tracklets. Each individual location in a
tracklet or path connecting tracklets becomes  a node of the player graph $G_p$,
it is then connected by an edge to the next location in the tracklet or path.

\subsection{Ball Graph}
\label{sec:ballGraph}

To  detect the  ball, we  use a  SVM~\cite{Hearst98} to  classify image  patches
in  each camera  view  based  on Histograms  of  Oriented  Gradients, HSV  color
histograms,  and motion  histograms.  We then  triangulate  these detections  to
generate candidate  3D locations and  perform non-maximum suppression  to remove
duplicates. We then  aggregate features from all camera view  for each remaining
candidate and train a second SVM to only retain the best.

Given  these high-confidence  detections, we  use  KSP tracker  to produce  ball
tracklets, as  we did  for people.  However, we  can no  longer use  the Viterbi
algorithm to connect them as the resulting connections may not obey the required
physical constraints. We  instead use an approach briefly  described below. More
details in supplementary materials.

\begin{figure}
    \includegraphics[width=\columnwidth]{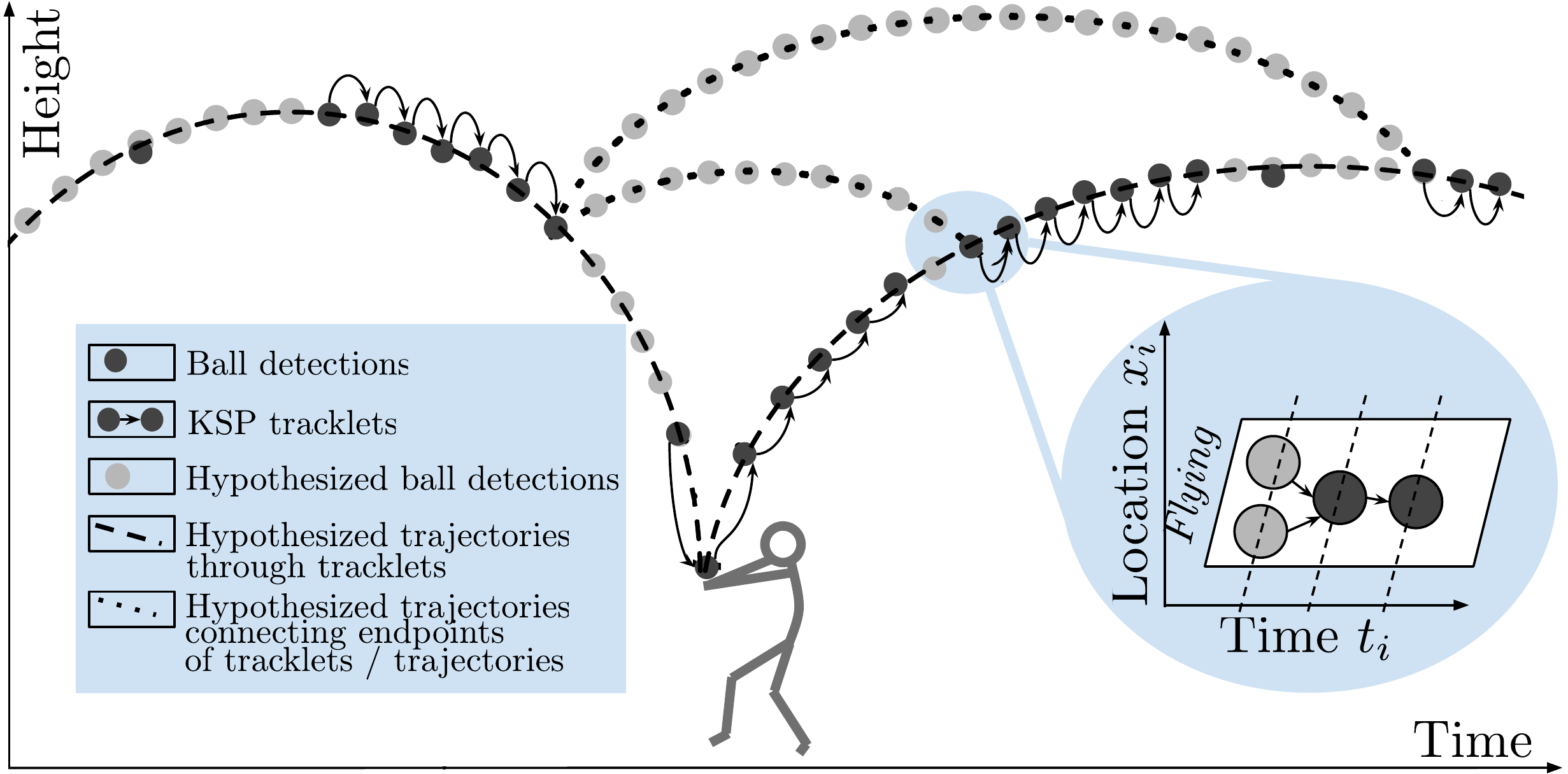}
\caption{An example of ball detections, hypothesized ball locations when it is missed, and graph construction.}
\label{fig:prunning}
\vspace{-0.5cm}
\end{figure}

To  model  the  ball  states  associated  to  a  physical  model,  we  grow  the
trajectories from each  tracklet based on the physical model,  and then join the
end points  of the  tracklets and  grown trajectories,  by fitting  the physical
model. An example of  such procedure is shown in  Fig.~\ref{fig:prunning}. To model
the state \textit{in\_possession}, we create a copy of each node and edge in the
players graph. To  model the state \textit{not\_present}, we create  one node in
each time  instant and connect  it to  the node in  the next time  instant, and
nodes for all other states in the vicinity of the tracking area border. Finally,
we add edges between  pairs of nodes with different states, as  long as they are
in the vicinity of each other (bold in Fig.~\ref{fig:factorGraph}(b)).

\comment{
For the  states associated  with physical model,  we grow  trajectories starting
from every pair of detections in every tracklet, and incorporating all following
detections that  fit the physical  model with the  maximum error of  $D_l$ (from
Eq.~\ref{eq:discreteCont}). We only  keep trajectories to which  maximal sets of
detections  fit. After  that,  we  join the  endpoints  of  tracklets and  newly
built  trajectories by  additional  trajectories that  fit  the physical  model.
Intuitively, first  step accounts for  missed detections,  and the second  - for
missed  trajectories,  as  shown  by  Fig.~\ref{fig:prunning}.  Each  trajectory
produces a path of nodes connected by edges in the ball tracking graph.}
\comment{
For  state \textit{in\_possession},  we  create a  copy of  each  node and  edge
in  the  players  graph.  \comment{As   detector  output,  we  use  the  largest
detection score  within the distance  of $D_p$  from the players  location (from
Eq.~\ref{eq:possConst}.)} For state \textit{not\_present}, we create one node in
each time  instant, and connect  it to  the node in  the next time  instant, and
nodes for all other states in the vicinity of the tracking area border. Finally,
we add edges between  pairs of nodes with different states, as  long as they are
in the vicinity of each other (bold in Fig.~\ref{fig:factorGraph},b).

Note that  while we generate  trajectories that obey physical  constraints, they
can not enforce a physical model - a ball can transit from one trajectory to
another in  the location in  a common node of  two. Structure of  ball tracking
graph  can  be  seen  as enforcing  physical  model  `locally',  while
second-order constraints  prevent the ball  from breaking the physical  model in
the ``intersections''.
}

\comment{ For the possession  state, for each node of the  people graph, 5 nodes
for the ball graph are created. These  nodes correspond to the state of the ball
being  possessed, and  are associated  with the  location that  projects to  the
player position  on the ground plane,  and has a height  of 50,100,\ldots,250cm.
For the absent ball state, we create only a single node in each time instance. %
Write  this  as an  actual  algorithm  \paragraph{Tracklet breakdown}  Tracklets
generated  by KSP  may  contain  some false  detections  and  not obey  physical
constraints. For  each tracklet  we generate maximally  long time  segments such
that all  points in  the time  segment lie  on the  trajectory specified  by the
constraints of the  ball state (parabola or  straight line) with an  error of at
most $D$. Detections in such segments become the new tracklets.

\paragraph{Tracklet extension} For each tracklet, we find the next (time-wise)
tracklet such that both tracklets can lie on the same trajectory with an error
of at most $D$. We continue extending tracklets as long as possible, and
afterwards extend the trajectory until it reaches the borders of the tracking
area. We only keep trajectories that include a maximum possible set of segments.

\paragraph{Trajectory generation} Additionally, we link endpoints of
tracklets and endpoints of the trajectories, for all pairs of tracklets and
endpoints inside a time window of a certain length. We experimentally found that
extending the length of the time window over 2 seconds does not bring benefits in
terms of better tracking, but becomes a heavier computational burden.

\paragraph{Detection association} For each node, we define the output of the
ball detector to be the largest output of the ball detector that fired within
the distance $D$ of this particular location.
}

\comment{
  \subsection{Obtaining solution}
It has been recently reported in~\cite{Kappes14} that integer
program solvers for structured energy minimization problems are competitive both
in terms of runtime and performance. We use Gurobi~\cite{Gurobi}
solver to obtain the solution or our problem. The result is obtained by minimizing
the gap between the lower bound of LP relaxation and upper bound of feasible
integer solutions. We require this gap to be below $1e-4$, obtaining the
solution close to the global optimum.
}

\comment{\subsection{Specific  states}   We  used   different  predefined   datasets  for
different   sports.   For   all   sports,  we   have   states   \textit{flying},
\textit{in\_possession}, \textit{not\_present}. For  volleyball, we additionally
have  state   \textit{strike}.  For  basketball,  we   additionally  have  state
\textit{pass}. For  football, we additionally have  state \textit{rolling}. More
details can be found in the supplementary materials.}

\comment{
\subsection{Post-processing} Our  approach does not encode  any domain knowledge
in the problem  formulation. While generating the videos  accompanying the paper
we used domain-specific post-processing for better appearance. Results we report
further  are  obtained without  post-processing,  and  for videos  the  tracking
accuracy was not  affected by more than 1\%. Details  are given in supplementary
materials.
}

\comment{: enforced linear,
rather then ballistic trajectory when the  ball is freely moving near the ground
in football, changed the state of the  ball from `possessed' to `flying' when it
was flying near the person for several frames but did not change the velocity in
volleyball, etc. We report the results without the post-processing in the paper.
The post-processing did not affect the  tracking accuracy more than 1\%. Details
of post-processing are given in supplementary materials.}

\comment{As  we  are   solving  our  problem  independently
of  the  sport,   we  do  not  encode  any  domain   knowledge  in  the  problem
formulation. However, after  having obtained the results,  we do domain-specific
post-processing. In particular, for volleyball we saw that the ball is sometimes
assumed to  be possessed by  the player it  flies nearby. We,  therefore, remove
possessions of  the ball that are  short and do  not change the velocity  of the
ball  significantly. \comment{Additionally,  while  we don't  bound  the Z  axis
acceleration of the  ball near the ground, in the  post-processing step we force
the ball  that bounces  of the  ground to have  only one  point of  contact with
the  ground.}  Details  about  specific  rules for  every  sport  can  be  found
in  the  supplemental materials.  We  report  results  with and  without  domain
post-processing.\amrmk{Post-processing only affects event  accuracy, by which we
are on  the top anyway. Should  we maybe just make  a small note saying  that we
apply  post-processing  when  generating  video  results,  and  detail  this  in
supplementary materials, for the sake of saving space?}}

\comment


\section{Experiments}
\label{sec:experiments}

In this  section, we compare  our results  to those of  several state-of-the-art
multi-view    ball-tracking    algorithms~\cite{Wang14a,Wang14b,Parisot15},    a
monocular one~\cite{Gomez14}, as well as  two tracking methods that could easily
be adapted for this purpose~\cite{Zhang12a,Berclaz11}.

\comment{
\pfrmk{This  sort of  contradicts  the introduction  statement  about all  other
state-of-the-art methods  being specific to a  particular sport. What do  we say
about those?} \amrmk{The~\cite{Parisot15,Gomez14} is sport-specific, at least it
is intended  for basketball. Other works  are used in multiple  sports, and this
makes sense  since we  are trying  to compare  to approaches  that are,  as our,
multisport. Second group, namely~\cite{Zhang12a,Berclaz11} are generic, they are
for object tracking,  not necessarily ball tracking. We compare  to them for the
purpose of showing that generic approaches don't work too well. Another point is
that many approaches from the related work don't have publicly available code.}}

We first describe  the datasets we use for evaluation  purposes. We then briefly
introduce the methods we compare against and finally present our results.

\subsection{Datasets}

We use  two volleyball, three  basketball, and  one soccer sequences,  which we
detail below.

\vspace{-3mm}
\paragraph{\basket{1}  and  \basket{2}}  comprise   a  4000-  and  a  3000-frame
basketball  sequences  captured   by  6  and  7   cameras,  respectively.  These
synchronized  25-frame-per-second  cameras  are  placed  around  the  court.  We
manually annotated each  10th frame of \basket{1} and 500  consecutive frames of
\basket{2} that feature flying ball, passed ball, possessed ball and ball out of
play.  We used  the  \basket{1} annotations  to train  our  classifiers and  the
\basket{2} ones to evaluate the quality of our results, and vice versa.

\vspace{-3mm}
\paragraph{\basket{APIDIS}}  is also  a basketball  dataset~\cite{De08} captured
by  seven  unsynchronized  22-frame-per-second  cameras.  A  pseudo-synchronized
25-frame-per-second version of the dataset is also available and this is what we
use. The  dataset is challenging because  the camera locations are  not good for
ball tracking  and lighting conditions  are difficult.  We use 1500  frames with
manually labeled ball  locations provided by~\cite{Parisot15} to  train the ball
detector,  and \basket{1}  sequence to  train  the state  classifier. We  report
our  results on  another 1500  frames  that were annotated  manually
in~\cite{De08}.

\vspace{-3mm}
\paragraph{\volley{1}  and  \volley{2}}  comprise  a 10000-  and  a  19500-frame
volleyball sequences captured by  three synchronized 60-frame-per-second cameras
placed at both ends of the court and  in the middle. Detecting the ball is often
difficult both because  on either side of the  court the ball can be  seen by at
most two cameras and because, after a strike,  the ball moves so fast that it is
blurred  in middle  camera  images.  We manually  labeled  each  third frame  in
1500-frame segments  of both sequences. As  before, we used one  for training and
the other for evaluation.

\vspace{-3mm}
\paragraph{\soccer{ISSIA}} is a soccer dataset~\cite{DOrazio09} captured by six
synchronized 25-frame-per-second cameras located on  both sides of the field. As
it is designed for  player tracking, the ball is often out of  the field of view
when flying.  We train  on the 1000 frames and report  results on another 1000. 
\\\\
In all  these datasets, the  apparent size  of the ball  is often so  small that
state-of-the-art monocular  object tracker~\cite{Zhang12a}  was never able  to track
the ball reliably for more than several seconds.
 
\subsection{Baselines}

We  use  several recent  multi-camera  ball  tracking algorithms  as  baselines.
To  ensure  a  fair  comparison,   we  ran  all  publicly  available  approaches
with\comment{,  or  asked their  authors  to  run  them \pfrmk{Have  to  explain
how  we  got  the Xinchao  results  using  an  algorithm  that is  not  online}}
the  same  set of  detections,  which  were produced  by  the  ball detector described in 
Sec.~\ref{sec:ballGraph}. We briefly describe these algorithms below. %

\begin{itemize}

  \vspace{-2mm}
  \item{\inter{}~\cite{Wang14b}} introduces  an Integer Programming  approach to
    tracking two types of interacting objects, one of which can contain another.
    Modeling the ball  as being ``contained'' by the player  in possession of it
    was  demonstrated  as  a   potential  application.   In~\cite{Wang15},  this
    approach   is   shown   to    outperform   several   multi-target   tracking
    approaches~\cite{Pirsiavash11,Leal-Taixe14} for ball tracking task.

  \vspace{-2mm}
 \item{\ransac{}~\cite{Parisot15}} focuses on  segmenting ballistic trajectories
 of the  ball and  was originally  proposed to track  it in  the \basket{APIDIS}
 dataset. Approach is  shown to  outperform the  earlier graph-based  filtering
 technique  of~\cite{Parisot11}.  We  found  that   it  also  performs  well  in
 our  volleyball datasets  that  feature many  ballistic  trajectories. For  the
 \soccer{ISSIA}  dataset, we  modified the  code to  produce linear  rather than
 ballistic trajectories.

  \vspace{-2mm}
 \item{\fos{}~\cite{Wang14a}} focuses  on modeling  the interaction  between the
   ball and the players, assuming that long passes are already segmented. In
     the  absence of  a publicly  available code,  we use the
     numbers   reported  in   the   article  for   \basket{1-2-APIDIS}  and   on
     \soccer{ISSIA}.  \comment{Authors  show  that their  approach  outperforms
     simple trajectory-growing approach.}   \comment{\pfrmk{That was really done
       manually?}  It uses information about the  players to decide which one is
     in possession.  \pf{The  authors report their results  on FIBAW-1, FIBAW-2,
       APIDIS and  ISSIA dataset.} \pfrmk{Do  we care since we  allegedly re-ran
       the  algorithm?}\amrmk{I am  sorry for  being misleading,  for FoS  I was
       planning to report the results from the original paper, since they are on
       the same  dataset. However, as  is shown in Xinchao's  thesis, InterTrack
       approach surpasses that  of FoS on several datasets.  Since  this is also
       an `internal' work, and we compare to a better work, should we maybe omit
       this approach? And when presenting  InterTrack, we could describe that it
       is better than several approaches.}}

  \vspace{-2mm}
   \item{\growth{}~\cite{Gomez14}}   greedily  grows   the   trajectories
   instantiated  from  points in  consecutive  frames.  Heuristics are  used  to
   terminate  trajectories,  extend  them  and link  neighbouring  ones.  It  is
   based  on the  approach of~\cite{Chen07} and  shown to outperform
   approaches based  on the Hough  transform. \comment{ that are  straight lines
   when projected onto  the ground plane. We reprojected our  detections from 3D
   to a single  view of each camera  to compare with other  methods. We selected
   the best results  among on the views.  } Unlike the other  approaches, it is
   monocular and we used as input  our 3D detections reprojected into the camera
   frame.

\end{itemize}
To refine our analysis and test the influence of specific element of our
approach, we also used the following approaches. 
\begin{itemize}

  \vspace{-2mm}
    \item{\maxdet{}.} To  demonstrate the importance of  tracking the ball,
    we give the  results obtained by simply choosing the  detection with maximum
    confidence.
    
  \vspace{-2mm}
    \item{\ksp{}~\cite{Berclaz11}.}  To demonstrate  the importance  of modeling
    interactions between the ball and the players, we use the publicly available
    KSP tracker  to track only the  ball, while ignoring the  players. \comment{
    only allowing the ball  to be in possession when a player  is nearby, we use
    the publicly  available KSP  tracker to  track separately  the ball  and the
    players. \pfrmk{I don't understand the previous sentence.}}
  
  \vspace{-2mm}
  \item{\ournp{}.}  To  demonstrate the  importance  of  physics-based
    second-order constraints of Eq.~\ref{eq:secondOrder}, we turn them off.

  \vspace{-2mm}
  \item{\ourns{}.} Similarly, to demonstrate the impact of keeping track of many
ball  states, we  assume  that  the ball  can  only be  in  one  of two  states,
possession and  free motion. 
\end{itemize} 

\comment{For free  movement, we generate ball  candidates by
extending  ball  trajectories  both  parabolically  and  linearly.  \pfrmk{Don't
understand  the last  sentence. Why  do you  need a  new generating  mechanism?}
\amrmk{When we had multiple states, we had a physical model associated with each
of them.  Therefore, for flying ball  all trajectories were parabolas.  Near the
ground all trajectories were straight lines. Here, since we don't have different
states, we  generate both  types of  trajectories for the  single state  of free
motion. This is not a new generating  mechanism, it is just putting together all
different trajectories  of different  kinds that  were previously  attributed to
different states.} 
\pfrmk{Shouldn't \ournp{} and FoS peform almost
the same? If  not, why not?} \amrmk{Did  you mean InterTrack? FoS  has little in
common with  our approach  at all.  As for InterTrack,  yes, it  performs mostly
similarly or better. When it performs better, it does so in cases where there is
a lot of  interaction and not much  free motion. In such cases  by not following
physically possible trajectory when transitioning between players, it is able to
recover faster.}}

\subsection{Metrics}

Our method tracks  the ball and estimates  its state. We use  a different metric
for each of these two tasks.

\vspace{-3mm}
\paragraph{Tracking  accuracy}  at  distance  $d$  is  defined  as  the  percent
of  frames  in  which   the  location  of  the  tracked  ball   is closer than
$d$ to the ground truth location.
\comment{More  formally:
\begin{small}
\begin{equation}
TA(d)={1  \over T}\sum^T\limits_{t=1}\mathbb{1}(||Pr(t)-Gt(t)||_2 \le  d).
\end{equation}
\end{small}}
The   curve   obtained   by   varying   $d$  is   known   as   the   ``precision
plot''~\cite{Babenko11}.   When the ball is 
\textit{in\_possession}, its location is assumed to be  that of  the player
possessing it.  \comment{\pf{We take $d$  to be  the Euclidean distance  in the
ground plane except in the two followings  cases.} If the ball is reported to be
\textit{in\_possession}  and really  is,  we take  $d$ to  be  the ground  plane
distance  between  the player  who  has  it and  the  one  who is  believed  to.
\pfrmk{What happens if the ball is reported \textit{in\_possession} and isn't?}}
If the ball is reported to be  \textit{not\_present} while it really is present, or vice
versa, the distance is taken to be infinite.

\vspace{-3mm}
\paragraph{Event accuracy} measures how well we  estimate the state of the ball.
We take  an \textbf{event} to be  a maximal sequence of  consecutive frames with
identical ball states. Two  events are said to match if there  are not more than
$5$ frames during which one occurs and not the other. Event accuracy then is
a symmetric measure  we obtain by counting recovered events  that matched ground
truth ones, as  well as the ground  truth ones that matched  the recovered ones,
normalized by dividing it by the number of events in both sequences.

\comment{plus the number  of events from ground truth that  match with recovered
evant, over the total number of events in both sequences. We measure this metric
only  for \our{},  \ournp{},  \ourns{},  and \inter{}.  To  make the  comparison
possible with the last two when they report the ball as being free, we apply our
state classifier to  get the specific state of the  ball (\eg \textit{flying} or
\textit{rolling}).}

\comment{For approaches that do not report the  ball state, we use the output of
our  classifier to  predict it.  We give  further details  in the  supplementary
material.\pfrmk{Does it make any sense to do this?}}

\subsection{Comparative Results}

 We now  compare  our approach  to  the  baselines in  terms  of the  above
  metrics. As mentioned in Sec.~\ref{eq:MIP}, we obtain the players trajectories
  by  first running  the  code  of ~\cite{Fleuret08a}  to  compute the  player's
  probabilities   of   presence   in   each  separate   fame   and   then   that
  of~\cite{Berclaz11} to  compute their trajectories.  We  first report accuracy
  results when these  are treated as being correct, which  amounts to fixing the
  $p_i^j$ in Eq.~\ref{eq:finalEq}, and show that our approach performs well.  We
  then perform joint optimization, which yields a further improvement.

\vspace{-3mm}\paragraph{Tracking and Event Accuracy.}

\comment{
\begin{figure*}[t]
\begin{center}
\hspace{-1pt}
\begin{tabular}{ccc}
   \hspace{-0.4cm}\includegraphics[trim={1.8cm 6.5cm 2.3cm 6.5cm},clip,width=0.34\textwidth]{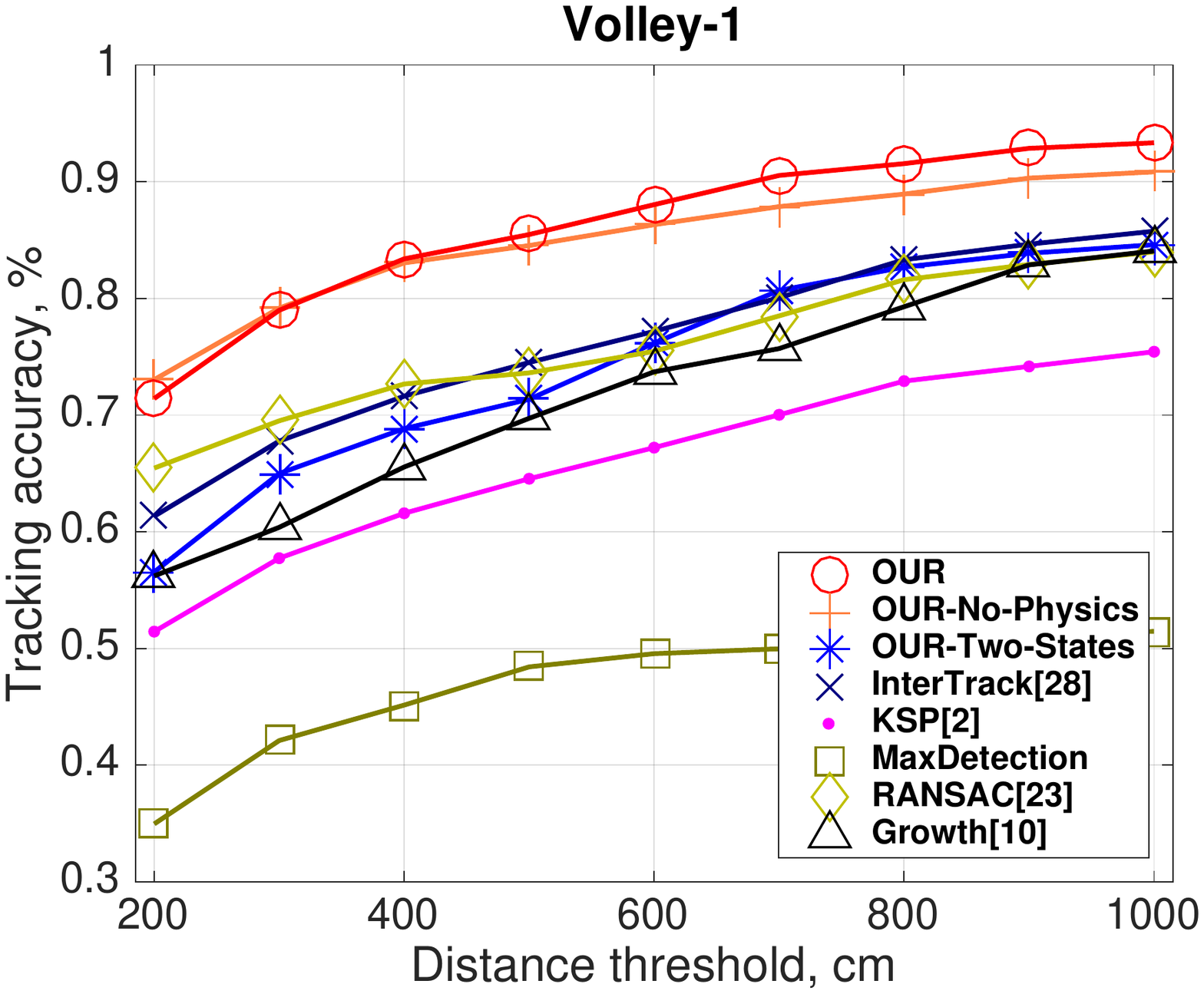} &
   \hspace{-0.4cm}\includegraphics[trim={1.8cm 6.5cm 2.3cm 6.5cm},clip,width=0.34\textwidth]{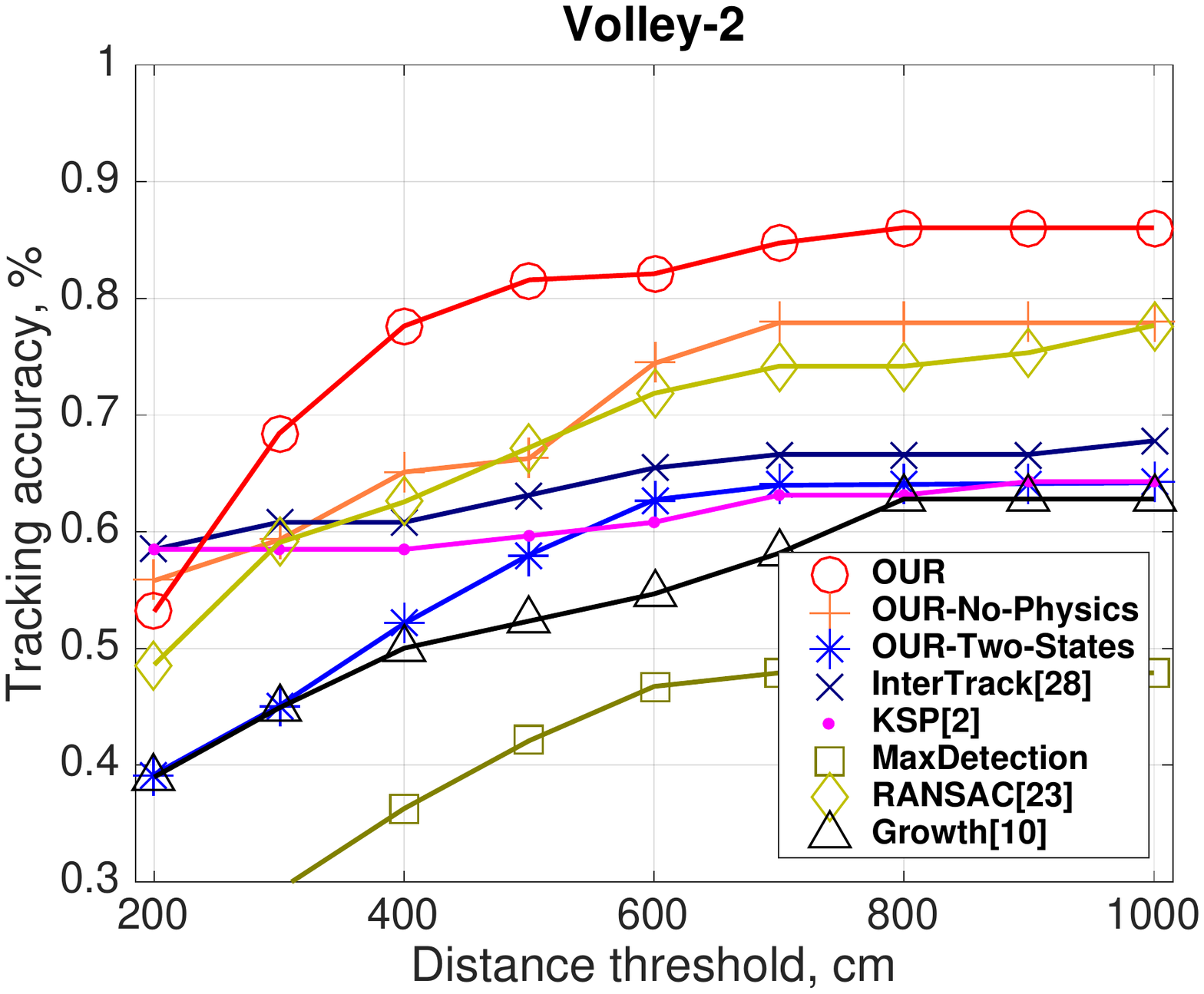} &
   \hspace{-0.4cm}\includegraphics[trim={1.8cm 6.5cm 2.3cm 6.5cm},clip,width=0.34\textwidth]{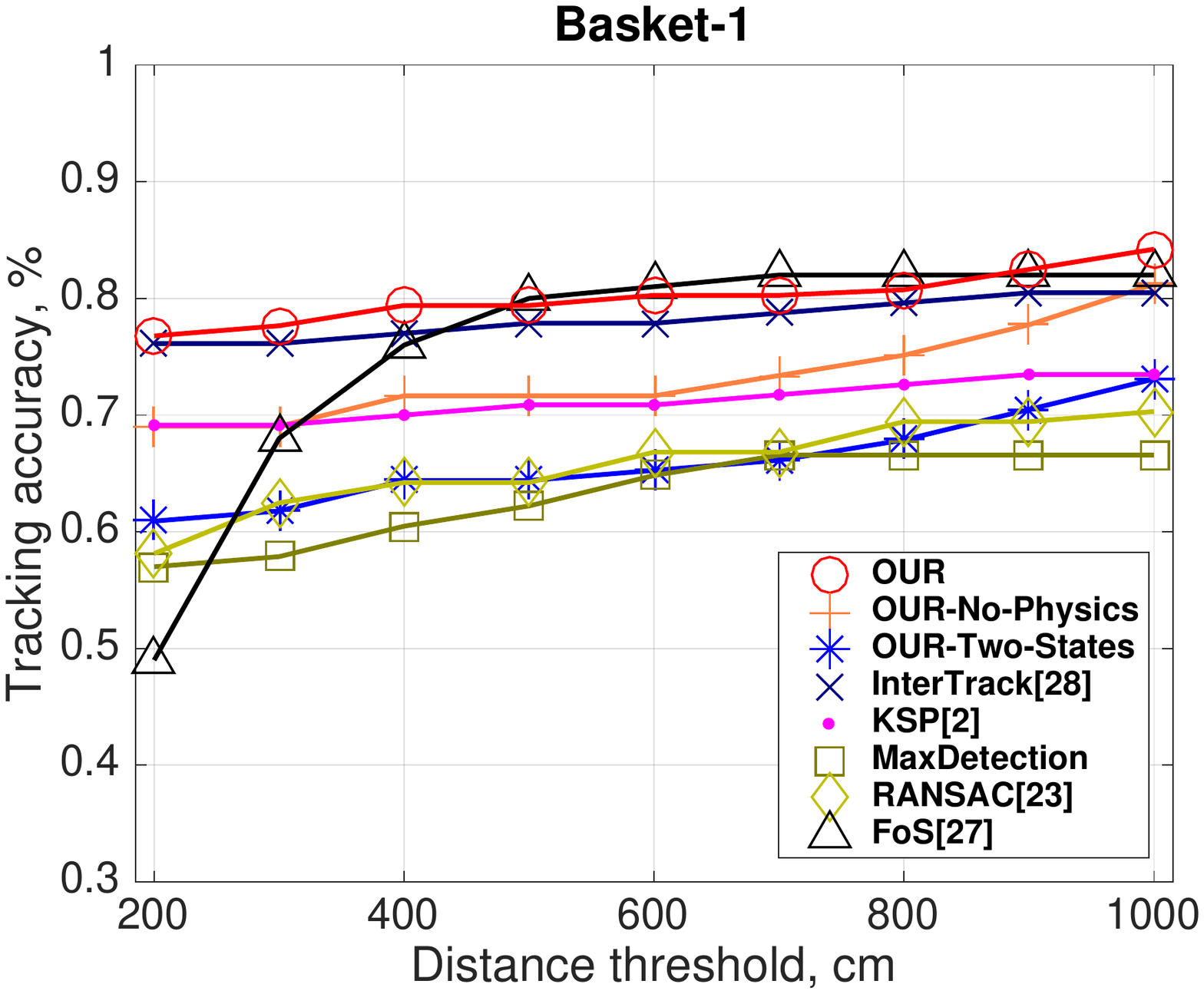}\\ 
	
   \hspace{-0.4cm}\includegraphics[trim={1.8cm 6.5cm 2.3cm 6.5cm},clip,width=0.34\textwidth]{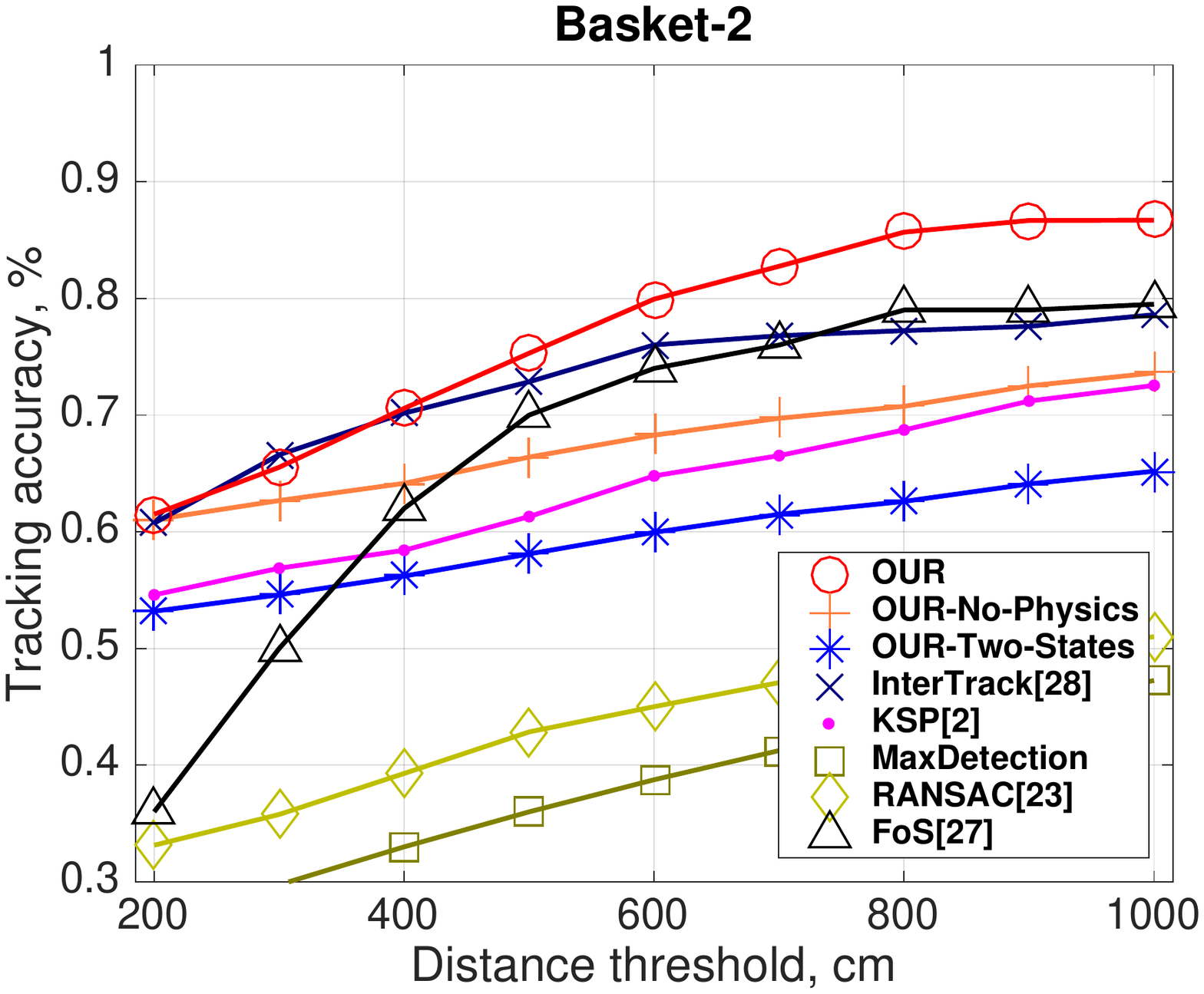} &
   \hspace{-0.4cm}\includegraphics[trim={1.8cm 6.5cm 2.3cm 6.5cm},clip,width=0.34\textwidth]{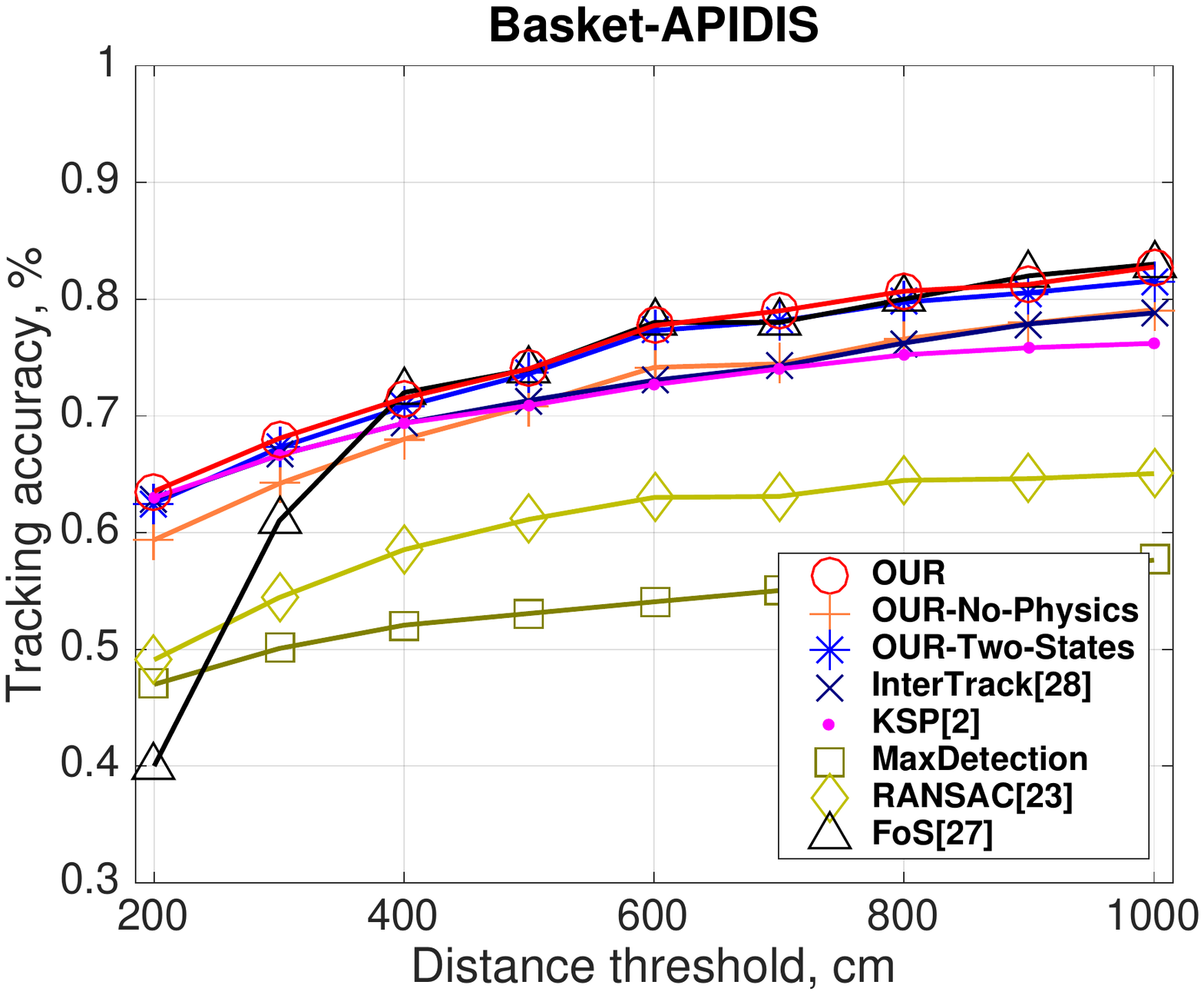} &
   \hspace{-0.4cm}\includegraphics[trim={1.8cm 6.5cm 2.3cm 6.5cm},clip,width=0.34\textwidth]{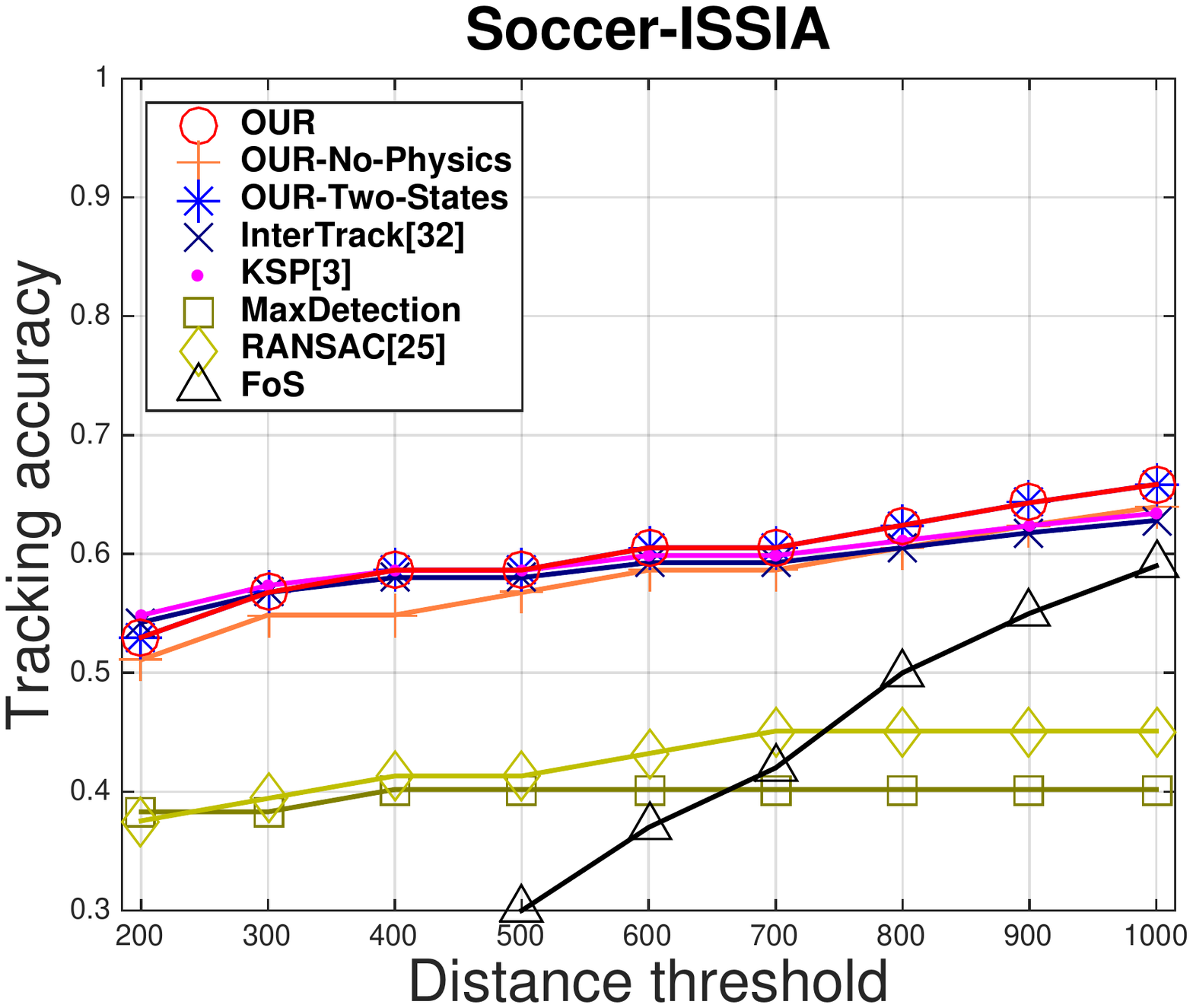}
   \\
\end{tabular}
\end{center}
\vspace{-3mm}
\label{fig:TAFigure}
\end{figure*}
}
\begin{figure*}
\begin{center}
\hspace{-1pt}
\begin{tabular}{cccc}
   \hspace{-0.4cm}\includegraphics[trim={1.6cm 6.5cm 2.3cm 6.5cm},clip,width=0.24\textwidth]{Figures/TA_Volley1.pdf} &
   \hspace{-0.4cm}\includegraphics[trim={1.6cm 6.5cm 2.3cm 6.5cm},clip,width=0.24\textwidth]{Figures/TA_Volley2.pdf} &
   \hspace{-0.4cm}\includegraphics[trim={1.6cm 6.5cm 2.3cm 6.5cm},clip,width=0.24\textwidth]{Figures/TA_Basket1.pdf} &	
   \hspace{-0.4cm}\includegraphics[trim={1.6cm 6.5cm 2.3cm 6.5cm},clip,width=0.24\textwidth]{Figures/TA_Basket2.pdf} \\ 

   \hspace{-0.2cm} (a) &
   \hspace{-0.2cm} (b) &
   \hspace{-0.2cm} (c) &
   \hspace{-0.2cm} (d) \\

   \hspace{-0.4cm}\includegraphics[trim={1.6cm 6.5cm 2.3cm 6.5cm},clip,width=0.24\textwidth]{Figures/TA_Apidis.pdf} &
   \hspace{-0.4cm}\includegraphics[trim={1.6cm 6.5cm 2.3cm 6.5cm},clip,width=0.24\textwidth]{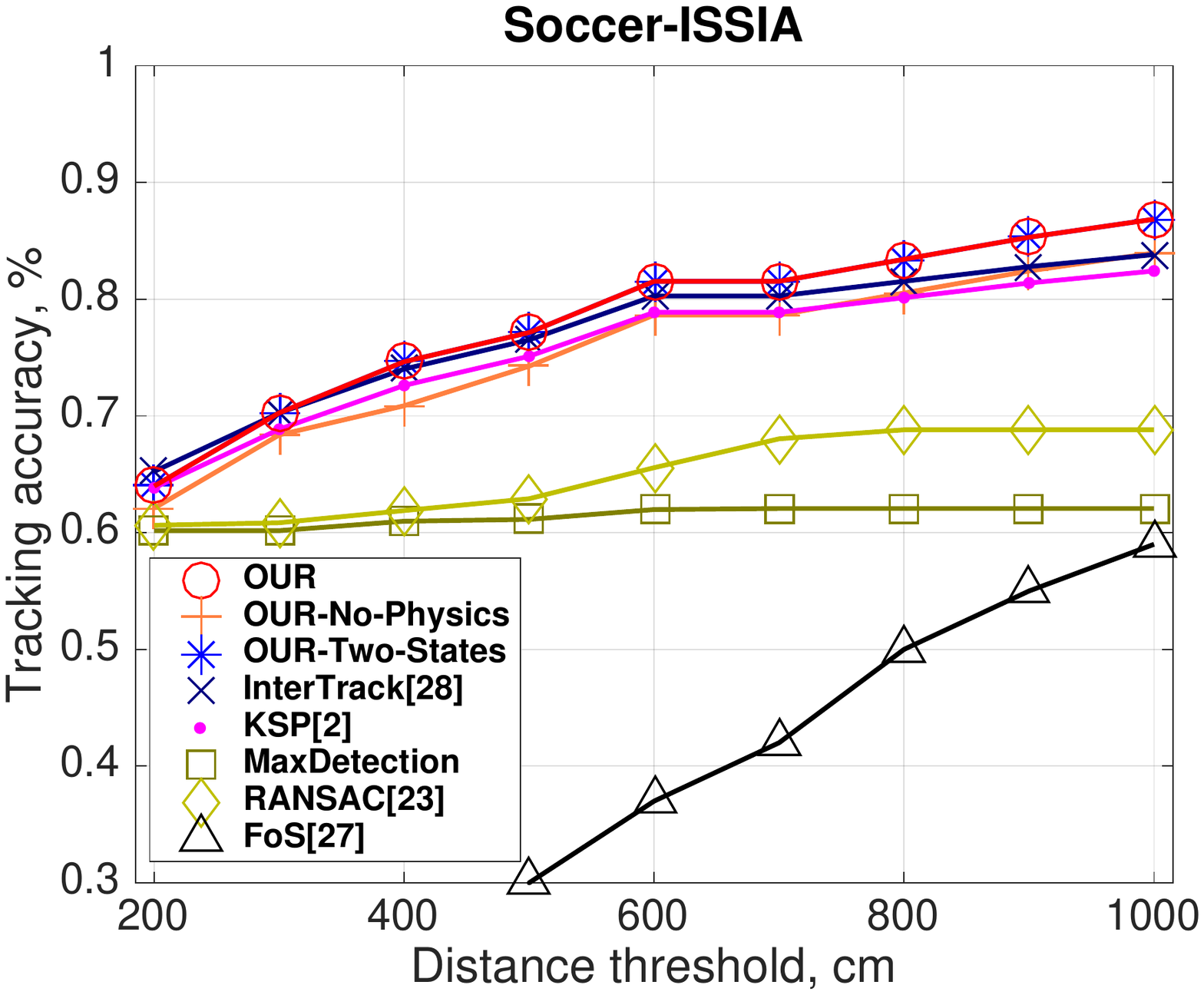} & \multicolumn{2}{c}{\includegraphics[trim={0 0 0 2cm},clip,width=0.48\textwidth]{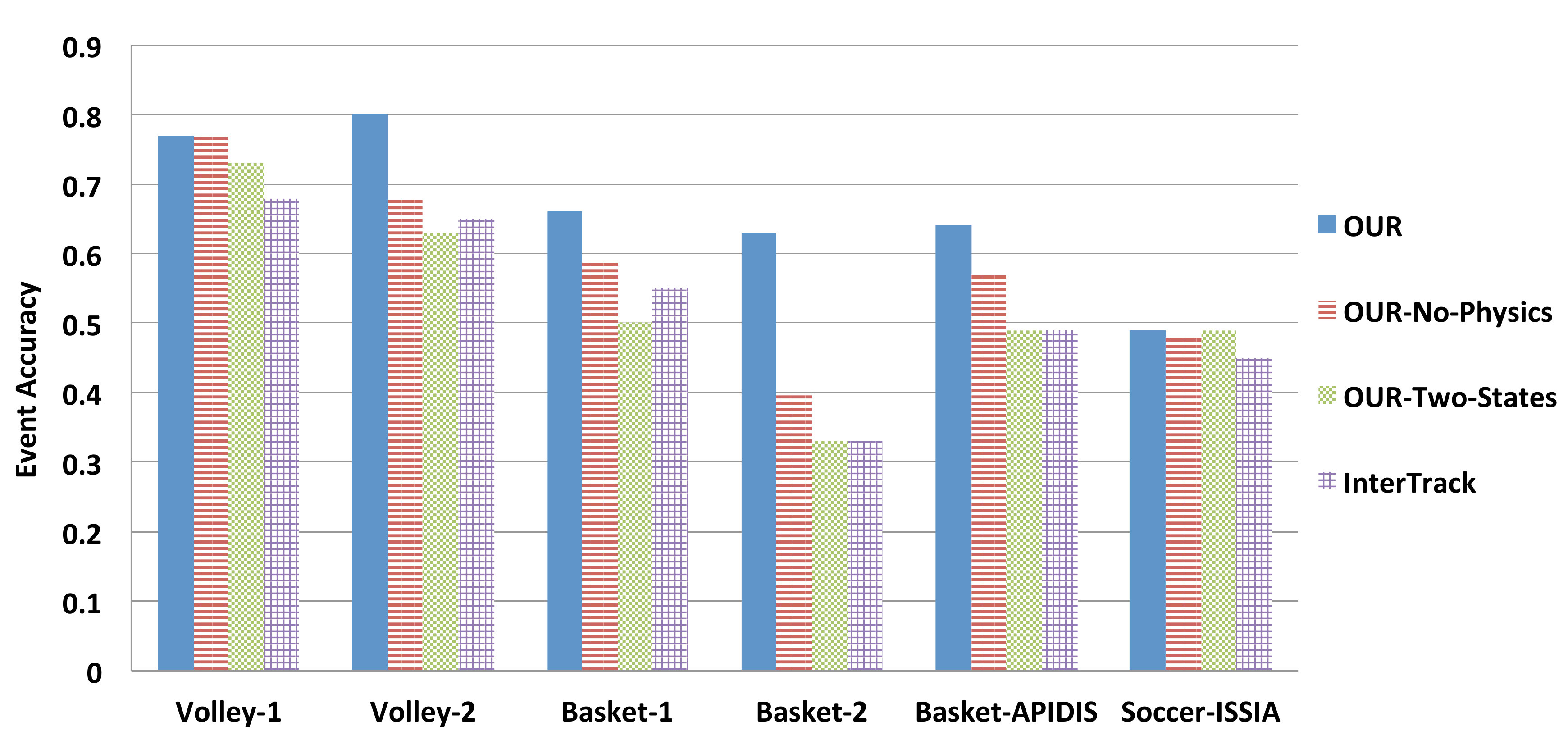}} \vspace{-0mm} \\
   \vspace{-5mm}
   \hspace{-0.2cm} (e) &
   \hspace{-0.2cm} (f) & ~~~~~~~~~~~~~~~~~~~~~~~~~~~~~~~~~~~~~~~~~~~~~(g) & \\
\end{tabular}
\end{center}
\vspace{-2mm}
\caption{Comparative  results.  {\bf  (a-f)} \our{}  outperforms the  other
approaches in  terms of ball accuracy,  followed by the other  methods that also
model ball/player interaction, \ournp{}, \inter{},  and \fos{} for larger values
of $d$. \comment{For \soccer{ISSIA}, results are reported during in-game moments
when  the ball  is visible. }{\bf(g)}  \our{} also  does best  in terms  of event
accuracy. }
\label{fig:TAFigure}
\end{figure*}

\begin{figure*}[t]
\begin{center}
\hspace{-1pt}
\begin{tabular}{ccc}
   \hspace{-0.2cm}\includegraphics[height=2.7cm]{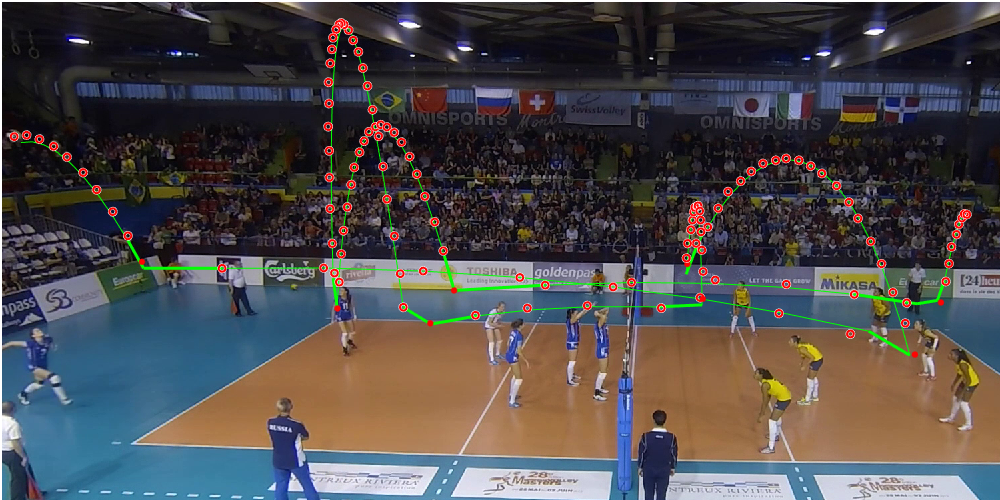} &
   \hspace{-0.2cm}\includegraphics[height=2.7cm]{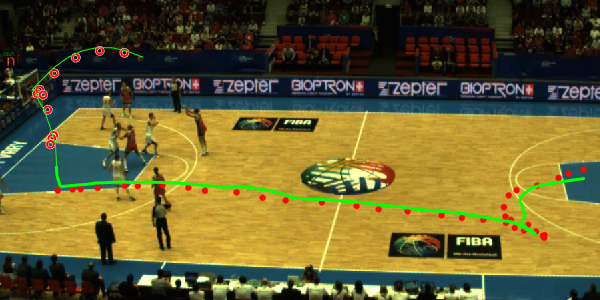} &
   \hspace{-0.2cm}\includegraphics[height=2.7cm]{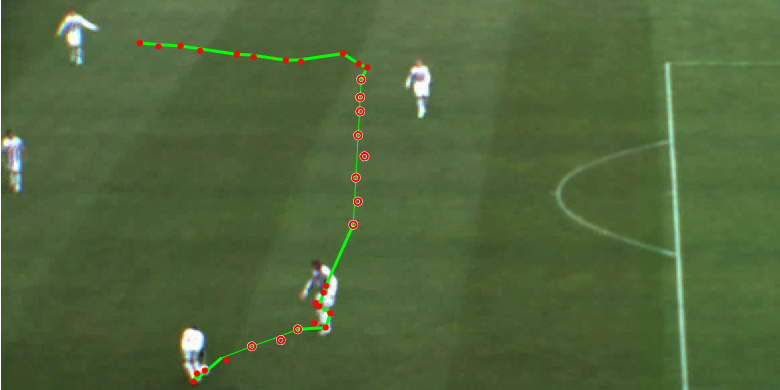} 
\\   
\hspace{-0.2cm} \volley{1} &
\hspace{-0.2cm} \basket{1} &
\hspace{-0.2cm} \soccer{ISSIA} 
\end{tabular}
\end{center}
\vspace{-6mm}

\caption{Visualisation of results on 3 10-second sequences from different sports. Cirlces indicate true ball location: empty circles correspond to free motion, filled circles indicate ball \textit{in\_possession}. Line indicates predicted ball locations: thick when predicted state is \textit{in\_possession}, thin otherwise. Best viewed in color.}

   \vspace{-2.0mm}
   \label{fig:results}
\end{figure*}

As shown in Fig.~\ref{fig:TAFigure}(a-f),  \our{} complete approach, outperforms
the  others on  all 6  datasets.  Two  other methods  that explicitly  model the
ball/player  interactions,  \ournp{}  and  \inter{},  come  next.   \fos{}  also
accounts for interactions but does  markedly worse for small distances, probably
due to the lack of an integrated second order model.

\vspace{-3mm}\subparagraph{Volleyball.}

The differences are particularly visible in the
Volleyball  datasets  that  feature  both  interactions  with  the  players  and
ballistic  trajectories.   Note that  \ourns{}  does  considerably worse,  which
highlights the importance of modeling the different states accurately.

\vspace{-3mm}\subparagraph{Basketball.}

The differences are  less obvious in the basketball datasets  where \ournp{} and
\inter{},  which  model the  ball/player  interactions  without imposing  global
physics-based constraints, also  do well.  This reflects the fact  that the ball
is handled much more  than in volleyball.  As a result,  our method's ability to
also impose  strong physics-based  constraints has less  overall impact.  

\vspace{-3mm}\subparagraph{Soccer.}

On the soccer dataset, the ball is only  present in about 75\% of the frames and
we report our results on those. Since  the ball is almost never seen flying, the
two  states   (\textit{in\_possession}  and  \textit{rolling})   suffice,  which
explains  the very  similar  performance  of \our{}  and  \ourns{}. \ksp{}  also
performs well because  in soccer occlusions during interactions  are less common
than in other sports. Therefore, handling them delivers less of a benefit.

\comment{On  the soccer  dataset, due  to camera  positioning, tracking  the ball  in
flight and during out of the game segments is very difficult, as it is often not
seen by  any camera.  We therefore  provide results on  the frames  without such
conditions.  We give  additional results  in supplementary  materials. Since  we
almost  never track  the flying  ball, two  states (\textit{in\_possession}  and
\textit{rolling}) are enough to explain the ball motions, which explains similar
performance of  \our{} and  \ourns{}. \ksp{}  also performs  well, as  in soccer
occlusions  during  interactions are  less  common  than  in other  sports,  and
explicitly handling them explicitly provides less improvement.}

Our method  also does  best in terms  of event accuracy, among the  methods that
report the state of the ball,  as shown in Fig.~\ref{fig:TAFigure}(g). As can be
seen in Fig.~\ref{fig:results}, both the  trajectory and the predicted state are
typically correct. Most state assignment errors  happen when the ball is briefly
assigned  to be  \textit{in\_possession}  of  a player  when  it actually  flies
nearby, or when  the ball is wrongly assumed  to be in free motion,  while is is
really \textit{in\_possession} but clearly visible.

\comment{

\am{   \ournp{}  also  works well,  as  in many  scenarios
enforcing local  physically viable trajectories  (by our graph  construction) is
already good enough. \inter{} approach effectively uses the interaction and also
obtains good results,  especially on basketball datasets, which  include lots of
interactions. The cost function of  this approach does not differentiate between
possession by different players, resulting in  some errors our approach does not
make.  \ransac{}  performs  best  on  volleyball dataset  due  to  abundance  of
parabolic trajectories,  but often fails  it situations with  many interactions.
\growth{}  performs  on par  or  worse  when it  can't  make  use of  detections
that  are not  in adjacent  frames. \ourns{}  performs equivalent  to \our{}  on
\soccer{ISSIA},  as tracking  of the  flying ball  was difficult  due to  camera
positioning, and  the presence  of third state  \textit{flying}, in  addition to
\textit{in\_possession}  and  \textit{rolling},  did  not  affect  the  results.
Reported \fos{} results are worse for  small tracking distances, probably due to
absence of integrated second order model.

}
}

\comment{
. We
noticed that it  often assigns possession to  the wrong player, due  to the fact
that its  cost function does not  have terms that correspond  to the possession.
Our solution, on the other hand, uses rare information about the ball detections
while  it  is  in  possession  to assign  the  possession  correctly.  \ransac{}
approach,  although originally  designed  for basketball  dataset, shows  better
results on  the volleyball dataset  due to abundance of  parabolic trajectories.
However, in  basketball setting, it  does not  allow reliable tracking  when the
ball is not shot towards the basket.  \comptrack{} is not able to track the ball
for more  than several dosen  frames and  therefore performs very  poorly, below
10\% in terms of  tracking accuracy. It is not shown on  the plots. For football
dataset,  results of  \ourns{} are  equivalent to  \our{}, as  positions of  the
cameras did  not allow  tracking the  flying ball,  or creating  the appropriate
training data,  and the  best results were  obtained by using  just 2  states of
\textit{in\_possession}  and \textit{rolling}.  \growth{} performs  on par  with
\ransac{} for \volley{1}, but in more difficult sequence of \volley{2} \ransac{}
is able  to make use  of detections not  in the neighbouring  frames, predicting
location  of the  ball  more often. }

\vspace{-3mm}\paragraph{Simultaneous tracking of the ball and players.}

  All the  results shown  above were  obtained by  processing sequences  of at
  least 500 frames.  In such sequences,  the people tracker is very reliable and
  makes few mistakes. This contributes to the quality of our results at the cost
  of an inevitable delay in producing  the results. Since this could be damaging
  in  the live-broadcast  situation,  we have  experimented  with using  shorter
  sequences. We show  here that simultaneously tracking the ball  and the players
  can mitigate the loss of reliability of  the people tracker, albeit to a small
  extent.

  \begin{table}[!h]
    \begin{center}
    \begin{tabular}{|c|c|c|}
        \hline
        Metric   & MODA~\cite{Kasturi09},\% & Tracking acc. @ 25 cm,\% \\
        \hline
        50       &  94.1 / 93.9 / 0.26 & 69.2 / 67.2 / 2.03 \\
        \hline
        75       &  94.5 / 94.2 / 0.31 & 71.4 / 69.4 / 2.03 \\
        \hline
        100      &  96.5 / 96.3 / 0.21 & 72.5 / 71.0 / 1.41 \\
        \hline
        150      &  97.2 / 97.1 / 0.09 & 73.8 / 73.0 / 0.82 \\
        \hline
        200      &  97.3 / 97.4 / 0.00 & 74.1 / 74.1 / 0.00 \\
        \hline
    \end{tabular}
    \end{center}
    \vspace{-0.3cm}
    \hspace{2.5cm} (a) \hspace{3cm} (b)
    \vspace{-0.3cm}
    \caption{Tracking  the ball  given the  players' locations  vs. simultaneous
      tracking  of the  ball and  players.  The  three numbers  in both  columns
      correspond to simultaneous  tracking of the players and  ball / sequential
      tracking of the players and then the ball / improvement, as function of the lengths
      of the sequences.  {\bf (a)} People tracking accuracy in terms of the MODA
      score.  {\bf (b)} Ball tracking accuracy.}
      \vspace{-0.6cm}
    \label{tab:simtrack}
\end{table}

As shown in Tab.~\ref{tab:simtrack} for the \volley{1} dataset, we need 200-long
frames to get  the best people tracking accuracy when  first tracking the people
by themselves first,  as we did before.  As the number of  frames decreases, the
people tracker becomes less reliable  but performing the tracking simultaneously
yields a small but noticeable improvement both for the ball and the players. The
case of Fig.~\ref{fig:motivation} is an example of this. We identified 3 similar
cases in 1500 frames of the volleyball sequence used for the experiment. 
\vspace{-0.3cm}

\comment{ Tracking the  players and the ball simultaneously  yields improvements over
tracking first  the players,  and the  tracking the  ball, assuming  the players
positions fixed. While  this improvement vanishes when the size  of the batch in
which we do the tracking is large, there are apparent benefits for smaller batch
sizes, as shown  by the Tab.~\ref{tab:simtrack}. Using smaller batch  sizes is a
necessary prerequisite for a near real-time performance system.}

\comment{ When  tracking was done on  small batches, we saw  an improvement both
for  people tracking  and ball  tracking  when we  tracked both  simultaneously,
rather  than  sequentially.   In  particular,  for  batches  of   70  frames  on
\volley{1}  dataset  we  observed  0.31\%  improvement  in  people  tracking  in
MODA~\cite{Kasturi09}  metric, in  situations  similar to  the  one depicted  in
Fig.~\ref{fig:motivation}. For  ball tracking we  observed a 2\%  improvement of
tracking accuracy  at the distance  of 25cm, as  well as better  event accuracy.
While similar results  were not observed on longer batches,  we consider this an
important result, as  in tracking systems using small batches  is often required
to obtain close to real-time performance.}

\comment{We have observed that people detectors are generally more reliable than
ball detectors,  and the graph  for people  tracking has approximately  order of
magnitude  less nodes  than the  graph  for ball  tracking, due  to huge  number
of  hypothesized  trajectories.  Presence  of  continous  variables  also  makes
optimization harder  for the ball. Solving  the problem for both  people and the
ball  virtually does  not  affect  the computation  time,  but sometimes  yields
improvements. When tracking is done in long enough batches (hundreds of frames),
we  have  seen no  improvement  in  the  people  tracking quality,  whether  the
problem  was solved  both for  ball and  players or  for players  only. However,
to  obtain close  to  real-time  performance using  small  batches is  required.
For  batches  of  size  70  on  \volley{1}  dataset  we  have  witnessed  0.31\%
improvement in  people tracking, in scenarios  similar to the ones  described in
Fig.~\ref{fig:motivation}.  For  the  ball  tracking, the  difference  was  more
significant,  both  because  errors  in trajectory  building  propagate  through
frames, and because there is only one ball, and reached 2\%.}

\comment{Given  long  enough sequences  of  frames,  we  have not  observed  any
improvement in ball  tracking in the scenario where ball  and people are tracked
together, compared  to the scenario  when they are tracked  separately. However,
for situations  where computation  time is an  issue and the  size of  the frame
batch in  which tracking  is done  is small  (\eg 50  frames), we  observed both
improvement  in  ball  and  people  tracking. One  such  scenario  is  shown  in
Fig.~\ref{fig:simTracking}. For  such short sequences  we saw an  improvement in
tracking accuracy at the level of 25cm by X\%, and improvement in event accuracy
at the margin of 5 frames by X\%.}

\comment{\paragraph{Continuous constraints and solution time}

As we observed, the time required to solve the program~\eqref{eq:mainFlow} is
significantly decreased compared to the version that does not introduce physical
constraints \eqref{eq:secondOrder}. For better performance, we propose a
relaxation of these constraints that require only integer variables, and
evaluate it in the terms of quality of the solution and the solution time. More
formally, we substite the continous variables $P^t$ in \eqref{eq:secondOrder} by
their definition in \eqref{eq:discreteCont}. By doing so, we allow the change of
acceleration at a margin of at most $D$, as defined in \eqref{eq:discreteCont}.
This does not enforce a trajectory with zero velocity change, but still prevents
rapid velocity changes. As Fig.~\ref{fig:solutionTime} shows, we see only a
slight decrease in the tracking and event accuracy, and the method still
outperforms all other approaches. Computation time, however, decreased to that
of the problem without the second order constraints.
}


\section{Conclusion}  We have introduced  an approach to ball  tracking and
state estimation  in team sports. It  uses Mixed Integer Program  that allows to
account for  second order motion  of the ball, interaction  of the ball  and the
players, and different  states that the ball can be  in, while ensuring globally
optimal solution.  We showed our  approach on several real-world  sequences from
multiple team sports. In  future, we would like to extend  this approach to more
complex tasks  of activity recognition and  event detection. For this  purpose, we
can treat events as  another kind of objects that can  be tracked through time,
and use interactions between events and other objects to define their state.

{\small
\bibliographystyle{ieee}
\bibliography{string,vision,learning,optim,misc}
}

\end{document}